%% file: article.tex
\documentclass{IEEEtran}
\IEEEoverridecommandlockouts
\usepackage{cite}
\usepackage{amsmath,amssymb,amsfonts}
\usepackage{xfrac}
\usepackage{algorithmic}
\usepackage{graphicx}
\usepackage{textcomp}
\usepackage{xcolor}
\usepackage{tikz}
\usepackage{subcaption}
\usetikzlibrary{positioning}
\usepackage{hyperref}
\usepackage{cleveref}
\usepackage{pgfplots}
\usepgfplotslibrary{groupplots}
\usepackage{tikzscale}
\pgfplotsset{compat = 1.9}

\usepackage{adjustbox}
\usepackage[ruled,vlined]{algorithm2e}
\usepackage{paralist}
\usepackage{tabularx,multirow}
\usepackage{booktabs}
\usepackage{bm}

\makeatletter
\newcommand{\removelatexerror}{\let\@latex@error\@gobble}
\makeatother

\SetKwInput{KwInit}{Init}

\definecolor{myblue}{rgb}{0.0, 0.44, 1.0}
\definecolor{myred}{rgb}{1.0, 0.0, 0.22}
\definecolor{mygreen}{rgb}{0.01, 0.75, 0.24}
\definecolor{mypurple}{rgb}{0.59, 0.44, 0.84}
\definecolor{myorange}{rgb}{1.0, 0.51, 0.26}

\newcommand{\norm}[1]{\left\lVert#1\right\rVert}

\newcommand{\featureExtractor}{f_{\theta}}
\newcommand{\baseClassifier}{c_{W_b}}
\newcommand{\novelClassifier}{c_{W_n}}

\def\BibTeX{{\rm B\kern-.05em{\sc i\kern-.025em b}\kern-.08em
    T\kern-.1667em\lower.7ex\hbox{E}\kern-.125emX}}
\begin{document}

\newtheorem{definition}{Definition} 
\newtheorem{theorem}{Theorem}
\newtheorem{lemma}{Lemma} 
\newtheorem{proposition}{Proposition}
\newtheorem{corollary}{Corollary}
\newcommand{\louis}[1]{\color{blue}#1 \color{black}}
\newcommand{\myriam}[1]{\color{mypurple}#1 \color{black}}
\newcommand{\X}{\mathcal{X}}
\newcommand{\Y}{\mathcal{Y}}
\newcommand{\C}{\mathcal{C}}
\newcommand{\N}{\mathcal{N}}
\newcommand{\Edges}{\mathcal{E}}
\newcommand{\E}{\mathbb{E}}
\newcommand{\R}{\mathbb{R}}
\newcommand{\F}{\mathcal{F}}
\newcommand{\Prob}{\mathbf{P}}

\title{Predicting the Accuracy of a Few-Shot Classifier}

\author{Myriam~Bontonou,~\IEEEmembership{Student Member,~IEEE,} Louis~B\'ethune, and~Vincent~Gripon,~\IEEEmembership{Senior Member,~IEEE}%
\thanks{M. Bontonou and V. Gripon are with the Electronics Department, IMT Atlantique, Brest, France, with the Lab-STICC, Brest, France and with MILA, Universit\'e de Montr\'eal, Montr\'eal, Canada (e-mail: myriam.bontonou@imt-atlantique.fr). V. Gripon is also with Universit\'e Cote d'Azur.}%
\thanks{L. B\'ethune is with \'Ecole Normale Sup\'erieure de Lyon, Lyon, France and also with MILA, Universit\'e de Montr\'eal, Montr\'eal, Canada.}}

\maketitle

\begin{abstract}
In the context of few-shot learning, one cannot measure the generalization ability of a trained classifier using validation sets, due to the small number of labeled samples. In this paper, we are interested in finding alternatives to answer the question: is my classifier generalizing well to previously unseen data? We first analyze the reasons for the variability of generalization performances. We then investigate the case of using transfer-based solutions, and consider three settings: i)~supervised where we only have access to a few labeled samples, ii)~semi-supervised where we have access to both a few labeled samples and a set of unlabeled samples and iii)~unsupervised where we only have access to unlabeled samples. For each setting, we propose reasonable measures that we empirically demonstrate to be correlated with the generalization ability of considered classifiers. We also show that these simple measures can be used to predict generalization up to a certain confidence. We conduct our experiments on standard few-shot vision datasets.

\end{abstract}


\section{Introduction}
In recent years, Artificial Intelligence algorithms, especially Deep Neural Networks (DNNs), have achieved outstanding performance in various domains such as vision~\cite{krizhevsky2012imagenet}, audio~\cite{aytar2016soundnet}, games~\cite{silver2017mastering} or natural language processing~\cite{bahdanau2015neural}. They are now applied in a wide range of fields including help in screening and diagnosis in medicine~\cite{burt2018deep}, object detection~\cite{zhao2019object}, user behavior study~\cite{ma2019learning} or even art restoration~\cite{gupta2019restoration}\dots

Designing a DNN consists in finding an adequate network architecture and training it to perform a given task using available labeled data gathered in a training set. In practice, there is a risk of overfitting, that is to say the model focuses on details of the training samples and is not able to generalize well to new ones. This is why the generalization abilities of trained models are often evaluated using another set of labeled samples called the validation set.

Problematically, stressing the generalization of a model using a validation set requires having access to a large quantity of labeled data. Yet annotating data typically costs money or even more inconvenient, in some cases, the acquisition of data is in itself costly. An extreme case is when one has only access to a few labeled samples, referred to as few-shot in the literature. In such a case, trained models are likely to cause overfitting due to the small diversity of training data. The non-accessibility to a validation set is thus even more critical.

The problem of few-shot learning has known many contributions in the past few years. The core idea of most methods consists in exploiting data or knowledge gathered for other tasks. For example, methods that rely on transferring knowledge usually train a DNN on a first task using a huge amount of data. Then, a latent representation obtained through this DNN is used as a feature extractor for the few-shot task to be solved. The problem is that features have been trained for another task. Consequently, novel classes can divide a base class, or to the contrary, be a union of base classes. Even worse, the overlap may be ill-defined. Consequently, according to data and tasks, performances can range from very bad to very good. In practice, a question of paramount importance is then to be able to guess in which case we are. 

In most few-shot learning works, the ability of a network to generalize to previously unseen inputs is not assessed because of the lack of a validation set. Often, hyperparameters are tuned using the test set, which is not a good practice since the test set should only be used after the solution has been designed to compare to other works.

In this work, we are interested in tackling the following problem: how can we estimate the generalization ability of a classifier in the context of transfer-based few-shot learning? We study three different settings related to few-shot: i) supervised, where we only have access to a few labeled samples, ii) semi-supervised, where we have access to both a few labeled samples and a set of unlabeled samples and iii) unsupervised, where we only have access to unlabeled samples. In all cases, we propose reasonable measures that we empirically demonstrate to be correlated with the generalization ability of trained classifiers. We also show that these simple metrics can be used to predict generalization up to a certain confidence.

The paper is organized as follows. In Section \ref{sec:background}, we introduce the formalism and methodologies of few-shot learning, and the tasks we aim to solve. Section \ref{sec:relatedwork} is dedicated to a review of the related work. In Section \ref{sec:casestudy}, we highlight the main issues arising in transfer-based few-shot learning, while in Section \ref{sec:methods}, we propose measures to predict the accuracy of a classifier trained with few labeled samples. Their efficiency is assessed in the experiments of Section \ref{sec:experiments}\footnote{Code at \url{https://github.com/mbonto/fewshot_generalization}}. A summary of our results is presented in Section \ref{sec:conclusion}.  
  
\begin{figure*}[ht]
    \centering
    \begin{subfigure}[b]{0.48\linewidth} 
    \includegraphics[width=1.\linewidth]{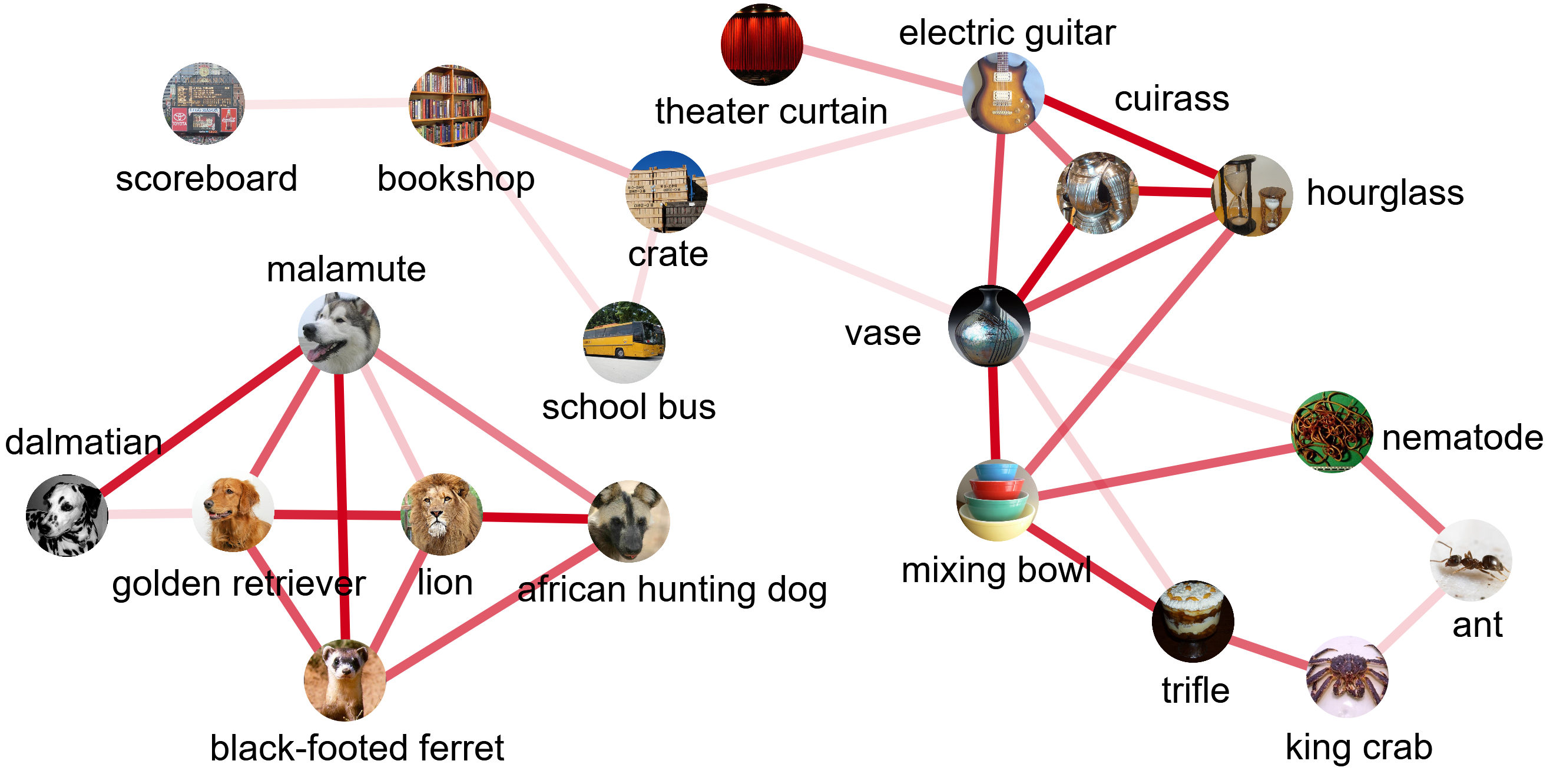}
    \caption{\textbf{densenet-m} backbone.}
    \label{fig:densenet}
    \vspace{1ex}
  \end{subfigure} 
  \begin{subfigure}[b]{0.48\linewidth} 
    \includegraphics[width=1.\linewidth]{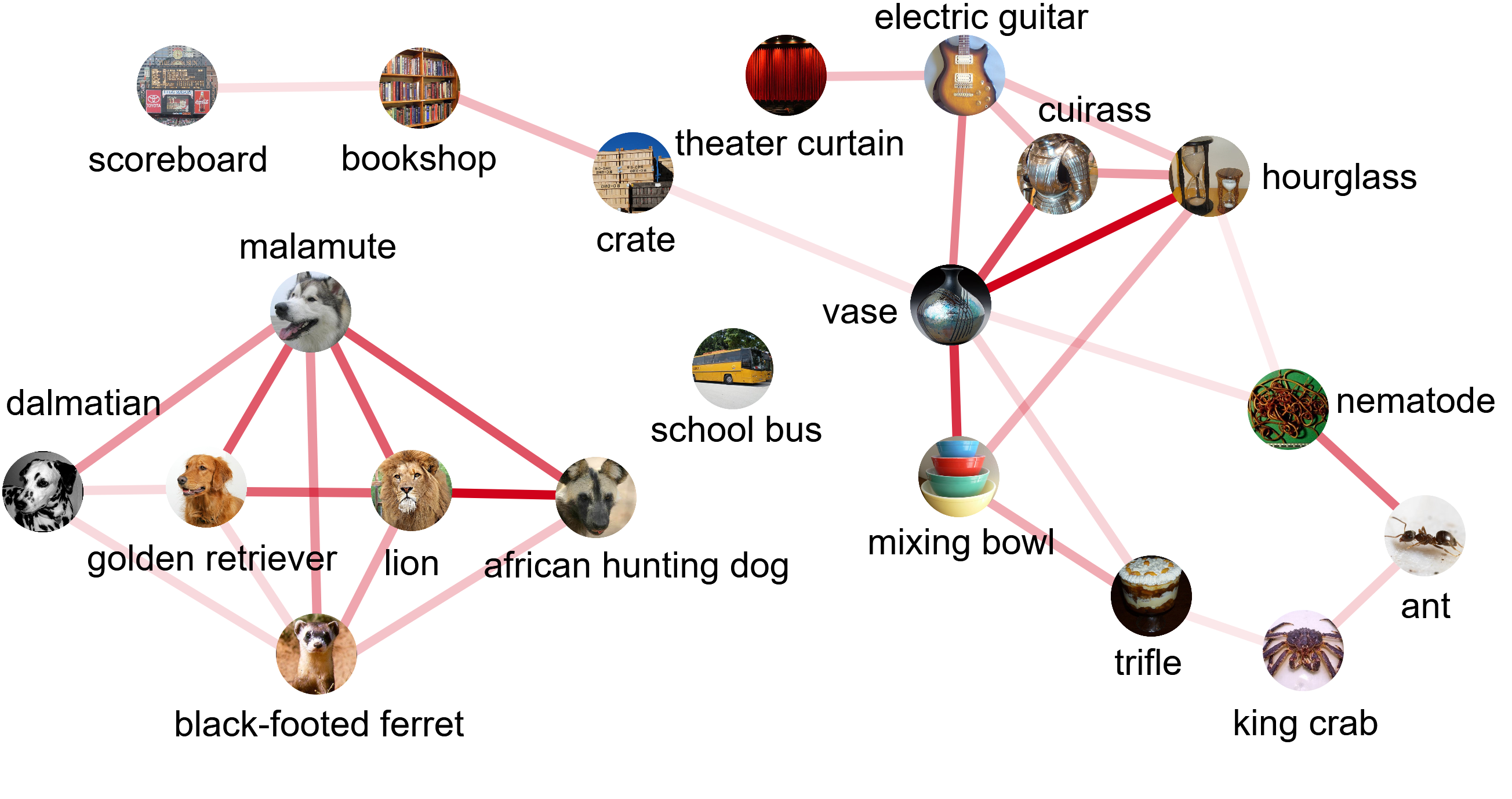}
    \caption{\textbf{wideresnet} backbone.}
    \label{fig:wideresnet}
    \vspace{1ex}
  \end{subfigure}
  \caption{Overlapping between classes of novel split, using the similarity measure introduced in Section~\ref{sec:casestudy_subsect2}. Position of vertices is arbitrary, and only the heavier edges are kept to ease reading (heavier edges are shown with a more intense color). Similarity of \textbf{wideresnet} with \textbf{densenet-m} is striking: the correlation of the weights of the edges on the complete graphs is 0.93.}
  \label{fig:similarity}
\end{figure*}

\section{Background}  \label{sec:background}
This section is intended as a reference for the rest of the article. It provides the formalism behind few-shot classification and ways of addressing it. The three settings we consider - supervised, semi-supervised and unsupervised, are detailed. The networks, classifiers and datasets used throughout the article, are also introduced.

Few-shot classification aims at learning to distinguish classes using only a few labeled samples. As such, there are not enough data to train a DNN from scratch.
In the literature, a usual approach is to use transfer learning, where a deep neural network is pretrained on a large available dataset, to be used as a feature extractor on the few-shot task. The large dataset is composed of what are called \emph{base classes} whereas the classes considered in the task are called \emph{novel classes}. In the remaining of this work, we denote $C_b$ the number of base classes and $C_n$ that of novel classes. Note that typically, it holds that $C_n \ll C_b$.

\subsection{Few-shot classification: a transfer-based approach}
The first step in transfer-based approaches consists in learning to extract feature representations from the abundant labeled data of the base classes. Then, a classifier is trained to associate the extracted representations of the few available labeled samples of the considered task with the corresponding novel classes. In the next paragraphs, we detail these steps.

\subsubsection{Backbone training on base classes}
Recall that we want to train a neural network using a large available dataset. This network is usually called \emph{backbone}. At the end of the training, we obtain a network function $g$, which can be written as ${g= \baseClassifier \circ \featureExtractor}$, where $\baseClassifier$ is a  $C_b$-way classifier whose parameters are $W_b$ and  $\featureExtractor$ is a convolutional Neural Network. Note that typically, $\baseClassifier$ is the last layer of the neural network, so that $\featureExtractor$ outputs the penultimate representation within the trained neural network when processing an input element. 

The features in base classes are a prior on the features expected to be found in novel classes. Hence, if the features are relevant for classification tasks over base classes, we expect them to be also relevant for the classification of novel classes.

\subsubsection{Feature Space}

Once the backbone has been trained, we can generate the feature representations of the data used in the few-shot task. Denote by $\mathbf{x}$ a sample we are given. We obtain $\mathbf{f} = \featureExtractor(\mathbf{x})$ its corresponding feature extraction. We call \emph{data space} the space where $\mathbf{x}$ has been drawn. Its extraction $\featureExtractor(\mathbf{x})$ is part of the \emph{feature space}. 

More formally, let $\X$ be the data space in which lay the images. The training set induces a distribution $\Prob_{\X}$ on this space. Let $\mathcal{F}$ denote the feature space (i.e the range of $\featureExtractor$). The feature extractor $\featureExtractor:\X\rightarrow\mathcal{F}$ induces a distribution $\Prob_{\mathcal{F}|\X}$.  

To solve the given task, the principle is to train a classifier to associate the extracted features with the corresponding labels. Denote by $\novelClassifier$ this classifier, then samples are classified using the composition: $\novelClassifier \circ \featureExtractor$, where $\novelClassifier$ is a $C_n$-way classifier.
  
Let $\Y$ denote the space of class labels, a finite set containing all possible labels for the task. The classifier on top of the backbone induces a distribution $\Prob_{\Y|\mathcal{F}}$. The predicted label associated with the features of a data sample $x \in \mathcal{F}$ is denoted $\tilde{y}_x$ while the true label is $y_x$.

\subsection{Studied problems}
We consider three different settings: a supervised setting, an unsupervised setting and a semi-supervised setting. The data space $\X$ carries very little usable information, so from now, all the examples are assumed to come from $\F$ with distribution $\Prob_{\mathcal{F}|\X}\Prob_{\X}$, through the use of the above-mentioned backbone. We now introduce the three problems in details.
  
\subsubsection{Supervised setting}
We consider $N$-way $K$-shot tasks. The number $N$ refers to the number of classes to discriminate from, and $K$ to the number of labeled samples we are given for each class. Denote by $\mathbf{x}_{ij}$ the $i$-th labeled element of class $j$, then the training set can be written as:
\begin{equation} \label{eq:supervised_def}
    S^{\text{supervised}}=\left\{(\mathbf{x}_{ij},y_j)| 1\leq i\leq K, 1\leq j \leq N \right\}\,.
\end{equation}

\subsubsection{Unsupervised setting}
We consider $N$-way $Q$-query tasks. We are given $Q$ unlabeled samples for each of the $N$ considered classes. The training set can be written as:
\begin{equation} \label{eq:unsupervised_def}
    S^{\text{unsupervised}}=\left\{(\mathbf{x}_{ij}, \bot)| 1\leq i\leq Q, 1 \leq j \leq N\right\}\,.
\end{equation}
Here, $\bot$ refers to the fact we do not know what the labels are.

\subsubsection{Semi-supervised setting}
This setting is the combination of the previous two. We consider $N$-way $K$-shot $Q$-query tasks. Some training samples are labeled, others are not. The training set is:
\begin{equation} \label{eq:semi_def}
    S^{\text{semi}}=S^{\text{supervised}}\cup S^{\text{unsupervised}}\,.
\end{equation}
This lastly considered case is often the most desirable, since we can benefit from both the unlabeled and labeled samples when training the classifier.

\subsubsection*{Remark}
In both unsupervised and semi-supervised settings, although we do not have access to the labels during training, we know that each class is equally present. This is the methodology considered in several research papers~\cite{hu2020exploiting, lichtenstein2020tafssl} for evaluation purposes. We also include, in Section~\ref{sec:experiments}, an additional experiment where the distribution of classes is unbalanced.  

\subsection{Experimental setting}

In this section, we introduce the considered datasets, backbones and classifiers.

\subsubsection{Datasets}
We consider two datasets. The first one is mini-ImageNet \cite{vinyals2016matching}. It has been generated from the bigger ImageNet database \cite{russakovsky2015imagenet}. It is split into $64$, $16$ and $20$ classes, in which $600$ images are available. The first split is used to train the backbone, the second to validate its generalization ability and the third one to generate the few-shot tasks. The second dataset is tiered-ImageNet \cite{ren2018meta}. The splits contain $351$, $97$ and $160$ classes, with roughly about $1000$ samples each. It is also extracted from ImageNet. The interest of tiered-ImageNet is that the semantic of classes has been studied with WordNet \cite{miller1995wordnet} to ensure that the considered splits contain semantically different classes. In both datasets, as in numerous studies \cite{wang2019simpleshot}, the images are resized to $84$ $\times$ $84$ pixels.

When generating a few-shot task, $N$ classes are uniformly drawn at random in the last introduced split. The $K$ and $Q$ samples to generate from each class are uniformly drawn without replacement. To assess the generalization performance in the supervised setting, we usually measure the classification accuracy that is achieved on $50$ samples uniformly drawn from the remaining items for each considered class (that is to say items that were not drawn to be part of the $K$-shot). In the unsupervised and semi-supervised settings, the performances are measured on the $Q$ samples per class.

\subsubsection{Backbones}
We consider three feature extractors. The first one is a Wide Residual Network \cite{BMVC2016_87} of 28 layers and width factor 10 described in \cite{mangla2020charting}. It has been trained on mini-ImageNet with a classification loss (classification error), an auxiliary loss (self-supervised loss) and fine-tuned using manifold mix-up \cite{pmlr-v97-verma19a}. Its results are among the best reported in the literature. In the following, we denote it by \textbf{wideresnet}. For more extensive benchmarks, we also use a DenseNet \cite{huang2017densely} trained on mini-ImageNet, and the same DenseNet trained on tiered-ImageNet. Both are described in \cite{wang2019simpleshot}. They are respectively referred to as \textbf{densenet-m} and \textbf{densenet-t} in the following. As advised in the original papers, we divide all feature vectors by their $\text{L}_2$-norm. Given $\mathbf{f}\in\mathcal{F}$, $\mathbf{f} \leftarrow \frac{\mathbf{f}}{\norm{\mathbf{f}}_2}\;.$

\subsubsection{Classifiers}
\paragraph{Supervised setting}
A classifier allowing nearly always to get the best performances in the supervised setting is the Logistic Regression (LR). Given $d$ the number of dimensions of the feature vectors, a weight matrix $\mathbf{W} \in \mathbb{R}^{d\times N}$, the matrix containing the features of all data samples $\mathbf{F} \in \mathbb{R}^{(NK)\times d}$, the output of the LR $\mathbf{P} \in \mathbb{R}^{(NK)\times N}$ is:

\begin{equation} \label{eq:LR}
\mathbf{P} = \text{softmax}(\mathbf{F}\mathbf{W})\,,
\end{equation}
where $\mathbf{p}[i,c]$ is the probability that the sample $i$ belongs to class $c$. Weights are learned using backpropagation to minimize a cross-entropy loss. It constrains the novel classes to be separated by hyperplanes in the feature space $\mathcal{F}$. A stronger classifier would allow the examples to lay on a manifold with more complex shapes, making clustering more challenging. If not specified otherwise, LR is trained on $50$ epochs with Adam optimizer \cite{kingma2014adam}, a learning rate of $0.01$ and a weight decay of $5e-6$.

\paragraph{Semi-supervised setting}
One of the state-of-the-art classifiers is an adapted Logistic Regression~\cite{hu2020exploiting}. The features extracted from a backbone are diffused through a cosine similarity graph $\mathcal{G}$ before being processed by a usual LR.

\vspace{0.25cm}
\begin{definition}[Cosine similarity] \label{def:cosine_similarity}
Given $\mathbf{f}_i$, $\mathbf{f}_j \in \mathcal{F}$, the cosine similarity function is defined as:
\begin{equation}
    \text{cos}(\mathbf{f}_i, \mathbf{f}_j) = \frac{\mathbf{f}_i^\intercal\mathbf{f}_j}{\norm{\mathbf{f}_i}_2\norm{\mathbf{f}_j}_2}\,.
\end{equation}
As in all backbones we use, the features are extracted after a ReLU function, all vectors in $\mathcal{F}$ contain non-negative values. Consequently, the output of the cosine similarity function ranges from $0$ to $1$.
\end{definition}
\vspace{0.25cm}

\begin{definition}[Cosine similarity graph] \label{def:cosine_simi_graph}
A cosine similarity graph $\mathcal{G} = \langle \mathcal{V}, \mathcal{E}, \mathbf{W}\rangle$ consists in a set of vertices $\mathcal{V}$ connected by a set of edges $\mathcal{E}$. The weights of the edges are stored in the adjacency matrix $\mathbf{W}$. Given two vertices $i$, $j$ and their feature vectors $\mathbf{f}_i$, $\mathbf{f}_j$,
\begin{equation}
\mathbf{W}[i,j] = \begin{cases}
                  \text{cos}(\mathbf{f}_i, \mathbf{f}_j)\;\text{if}\; \{i,j\} \notin \mathcal{E}\\
                  0 \;\text{otherwise}
                  \end{cases}\,.
\end{equation}
\end{definition}

We consider a cosine similarity graph $\mathcal{G} = \langle \mathcal{V}, \mathcal{E}, \mathbf{W}\rangle$ (see Definition~\ref{def:cosine_similarity} and~\ref{def:cosine_simi_graph}). After removing self-loops, we only keep the $k$-th largest values on each row and we normalize the resulting matrix as follows:
\begin{equation}
    \mathbf{E} = \mathbf{D}^{-\frac{1}{2}}\mathbf{W}\mathbf{D}^{-\frac{1}{2}}\,,
\end{equation}
where $\mathbf{D}$ is the diagonal degree matrix defined as:
\begin{equation}
    \mathbf{D}[i,i] = \sum_j{\mathbf{W}[i,j]}\,.
\end{equation}
Given $\mathbf{F} \in \mathbb{R}^{(NK+NQ)\times d}$ the matrix containing the features of all data samples and $\mathbf{I}$ the identity matrix, the new features are obtained by propagating the extracted features as follows:
\begin{equation}
    \mathbf{F}_{\text{diffused}} = (\alpha\mathbf{I} + \mathbf{E})^\kappa \mathbf{F}\,.
\end{equation}
In the reference paper~\cite{hu2020exploiting}, $\alpha$, $\kappa$ and $k$ are hyperparameters. The best ones found on $5$-way $5$-shot $15$-query tasks on mini-ImageNet are $\alpha=0.75$, $k=15$ and $\kappa=1$. We use these hyperparameters in all our experiments.

Finally, a LR is trained on the diffused features $\mathbf{F}_{\text{diffused}}$, with the same parameters as in the supervised setting.

\paragraph{Unsupervised setting}
The unsupervised setting is less studied in the few-shot literature. We hypothesize that, when features are well adapted to a $N$-way task, each class is associated with a cluster in the feature space. In that case, a standard clustering method consists in using a $N$-means algorithm. We use the implementation of the algorithm with default values as implemented in scikit-learn \cite{scikit-learn}. In order to compare results with the semi-supervised setting in a fair manner, we also propagate the features extracted from backbones through a cosine similarity graph as we explained it in the semi-supervised setting.
To evaluate the quality of the clustering, we compute its Adjusted Rand Index (ARI). This index ranges from $0$ to $1$, $1$ meaning that the data samples are exactly clustered according to their labels, and $0$ that the clustering is at chance level.

In the next section, we discuss related work and introduce ours in a broader context.

\section{Related Work} \label{sec:relatedwork}
In machine learning, there is an increasing interest for learning a task with few samples. The literature on this subject is referred to as few-shot learning, where the authors can mean either the supervised setting or the semi-supervised setting is considered. In this section, we detail some of the most important articles of this field.  

\subsection{Few-shot learning}
As training a Deep Neural Network (DNN) on few data samples from scratch typically leads to overfitting, several learning strategies have been developed over the past years. All these strategies share the idea of building a general-purpose representation (using a \emph{backbone} network) of data samples. For instance, in image classification, some knowledge is retrieved from a rich database containing training classes, called base classes. Then, this knowledge is transferred to perform a new task on novel classes.  

\paragraph{With meta-learning}
A first group of strategies uses meta-learning. It consists in using entire tasks as training examples.
Some \emph{optimization-based} works propose to learn a good initialization of the weights of the DNN over several training tasks, so that a new task can be learned with only a few gradient steps~\cite{finn2017model, rusu2018metalearning}.
In \emph{metric-based} works~\cite{snell2017prototypical, vinyals2016matching, sung2018learning, oreshkin2018tadam, ye2018learning}, the idea is to learn to embed the data samples in a space where classes are easily separable. Thus, once a new task occurs, the features of the novel samples are projected into this embedding (without any learning) and a simple classifier is trained to recognize the novel classes from these features. As the number of parameters to learn is reduced, the risk of overfitting is lower. There are many variants in the literature. For instance, in~\cite{snell2017prototypical}, they assume that it exists an embedding space where each class is represented by one point. Thus, a DNN is trained over several training tasks to work with a distance-based classifier, in which each class is represented by the average of its projected data samples. When a new task comes, the representations of the samples are extracted from the DNN, and the labels of the query samples are attributed according to the class of the closest representative.

\paragraph{Without meta-learning}
In a recent line of work, some methods do not focus on learning a relevant embedding from training tasks but on learning a relevant embedding from a single classification task involving all training classes at once~\cite{chen2018a, wang2019simpleshot, tian2020rethinking, mangla2020charting}. First, a DNN is trained to minimize a classification loss on base classes. A regularization term such as self-supervision~\cite{mangla2020charting, tian2020rethinking} or Manifold Mixup~\cite{mangla2020charting} is sometimes added to the loss to learn more robust features. Then, the features of the samples of the few-shot task are extracted from the DNN (often using the features of its penultimate layer, just before the classifier). Finally, a simple classifier, such as a Nearest Class Mean~\cite{wang2019simpleshot} or a Cosine Classifier~\cite{mangla2020charting}, is trained on the extracted features to distinguish between classes.
In~\cite{wang2019simpleshot}, the authors show that simple transformations, such as normalizing extracted features with their $\text{L}_2$-norm, help the classifier generalizing better.
Using self-supervision and Manifold Mixup, the article~\cite{mangla2020charting} achieves state-of-the-art performances on benchmark datasets. That is why, in this article, we allow to restrict our study to few-shot learning solutions based on pretrained feature extractors.  

\subsection{Generalization to new classes}
In transfer-based few-shot learning, the challenge is to learn representations on training classes, which are suitable for novel classes. Indeed, the generalization ability of a classifier is linked to the distribution of the representatives of the data samples in the feature space. However, it is not easy to estimate whether the learned embedding space suits novel classes well.
\paragraph{Learning diverse visual features}
The generalization ability of the classifiers depends on the relevance of the extracted features for a new task. Inspired by works in deep metric learning, the authors of~\cite{milbich2020diva} propose to learn representations capturing general aspects of data. They optimize a DNN to perform a range of tasks enhancing class-discriminative, class-shared, intra-class and sample-specific features. Although they do not apply their method to few-shot tasks, it could help improving the generalization. Similarly, self-supervised learning and Manifold Mixup used in~\cite{mangla2020charting} improve the accuracy on few-shot tasks. Another way to learn richer representations is to use additional unlabeled samples. 

\paragraph{Using additional unlabeled data samples}
When unlabeled samples are available, they can be used to infer more adapted representations of data samples to distinguish between novel classes. In the literature, two settings are studied. In one of them, the unlabeled samples are the query samples on which the accuracy of the classifier is evaluated. This is the setting we consider in this article. In the other setting, the unlabeled samples are just additional samples and they are not used to test the classifier.
The authors of~\cite{lichtenstein2020tafssl} consider both settings. They look for a linear projection which maximizes the probability of being in the correct class. More precisely, an unsupervised low-dimensional projection (PCA or ICA) is first applied on the features to reduce their noise. Then, data samples are clustered using two possible methods: a Bayesian $K$-Means or a Mean-Shift approach followed by a NCM classifier. In~\cite{hu2020exploiting}, the features of the data samples are diffused though a similarity graph computed from the few-shot samples and from the unlabeled samples before being used in a classifier. As these works use additional information, the generalization performances are increased.

\paragraph{Learning good representations}  
Learning efficient representations has always been a concern for deep learning~\cite{bengio2013representation}. Invariant Risk Minimization \cite{arjovsky2019invariant} and $\nu$-Information \cite{xu2020theory} have been proposed as theoretical frameworks to detail the properties a good representation should exhibit when connected to a (mostly linear) classifier. Other works focus on maximization of Mutual Information (following InfoMax principle) such as Deep Infomax~\cite{hjelm2018learning}. Losses (like in~\cite{milbich2020diva} or in~\cite{wang2019domain}) are designed to enforce some geometry in latent space based on similarity measures (such as sharing the same label). 
Robust few-shot learning for user-provided data~\cite{lu2020robust} is proposed to handle outliers within training samples.

\paragraph{Evaluating the generalization ability}
The generalization ability of a few-shot solution can be improved by designing more relevant representations of data. However, this ability is ill-evaluated. In standard deep learning, the generalization performance is computed on a validation set. Here, we do not have enough samples to afford such a procedure. Thus, the question of interest in this study, which has not been handled so far in the literature, is not how to improve the generalization performance but really how to evaluate it.

\section{Case study on mini-ImageNet} \label{sec:casestudy}

The current case study section has been designed to outline how difficult it is to predict the generalization ability of a classifier on a few-shot task.
We begin with a motivating observation. With mini-ImageNet, we can generate many $5$-way $5$-shot tasks. As mini-ImageNet contains hundreds of samples per class, we also generate test sets for all tasks. Solving them using a combination of \textbf{wideresnet} with LR, we observe that the performances on the test sets vary from $55.6\%$ to $98\%$ on a sample of $10,000$ runs. 

To better understand the reasons for these variations, we conduct an experiment, detailed in subsection~\ref{sec:casestudy_subsect1}. In brief, the accuracy on a task mainly depends on two variables, the distribution of classes with respect to each other and the random selection of labeled samples within classes. On mini-ImageNet, we show that the impact of the choice of shots is minor compared to the one of the classes. In subsection~\ref{sec:casestudy_subsect2}, we propose an interpretation of the difficulty induced by the choice of classes.

\subsection{Identifying the source of overfitting} \label{sec:casestudy_subsect1}
In order to look for the origin of the difficulty of a few-shot task, we propose the following experiment on the mini-ImageNet dataset. We consider $5$-way $5$-shot tasks and we extract the features of the samples with \textbf{wideresnet}. When we generate a task, we fix either the choice of classes or the choice of shots within classes. While varying the other variable, we look at the standard deviation of the accuracy of the task. If the standard deviation is low, we consider the variable has little impact.

\begin{figure}[t]
\input{Pseudo-codes/Random}
\input{Pseudo-codes/FixedClasses}
\input{Pseudo-codes/FixedShots}
\end{figure}

First, we compute the standard deviation of the accuracy over random tasks (see Algorithm~\ref{alg:random}). Second, we explore the impact of the choice of shots within classes on the accuracy of a task. We compute the standard deviations of the accuracy while fixing the classes (see Algorithm~\ref{alg:fixedclass}). Third, we explore the impact of the choice of classes by computing the standard deviations of the accuracy while fixing the shots within each class (see Algorithm~\ref{alg:fixedshot}).

The standard deviation of the accuracy over random $5$-way $5$-shot tasks is $5.85\%$. The average of the standard deviations over tasks with fixed classes and random shots is $2.02\%$. Its standard deviation is $0.35\%$. The average of the standard deviations over tasks with random classes and fixed shots is $4.37\%$. Its standard deviation is $0.61\%$. We observe a higher variability within the choice of classes than within the selection of labeled samples within classes. 
\begin{figure*}[!ht]
    \centering
    \includegraphics[width=16cm,height=29cm,keepaspectratio]{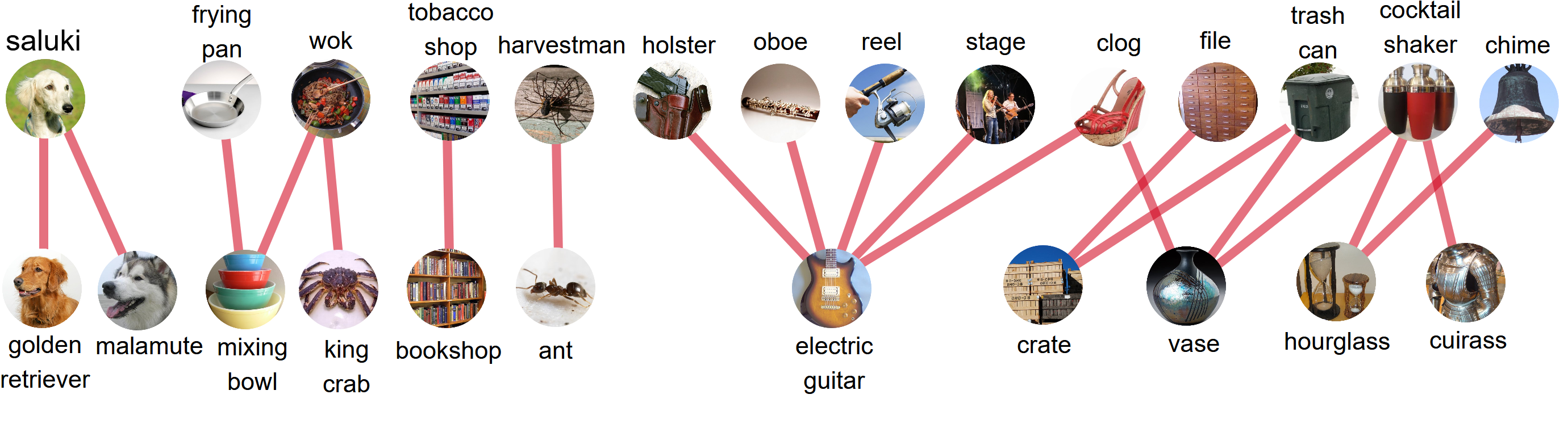}
    \caption{Confusion between \emph{base} classes (top) and \emph{novel} classes (bottom) with the \textbf{densenet-m} backbone. Position of vertices is arbitrary. Only the heavier edges have been kept to ease reading. Details in Section~\ref{sec:basenovelconfusion}.}
    \label{fig:basenovelconfusion}
\end{figure*}

In this subsection, we considered both variables independent. However, in reality, they are not. If we consider the closest classes, the choice of shots is much more important. The average of the standard deviations over tasks with fixed classes and random shots is 2.26\%. Its standard deviation is 0.38\%. The average of the standard deviations over tasks with random classes and fixed shots is 3.97\%. Its standard deviation is 0.31\%. Similarly, if classes are far away, the choice of shots does not have an influence anymore.

In the next section, we propose to unveil the inherent difficulty of some tasks. To understand the impact of the choice of classes on the generalization, we propose to study the distribution of classes in the input space of the classifier (which we called the feature space). Thus, we look at the topological space of data samples after they went through the feature extractor.

\subsection{Measuring class confusion}
\label{sec:casestudy_subsect2}

We show in the previous subsection that, for a fixed backbone, the variability in accuracy seems to be more dependent on the choice of classes than on the choice of examples inside those classes. Then arises the question: \emph{what are the pairs of classes $\{A,B\}$ hard to distinguish for a given backbone?}  
  
We refer to those pairs of classes as \emph{hard tasks}. The natural follow up being: \emph{are the hard tasks the same for different backbones?}

\subsubsection{Graph based measure of domain overlapping}
  
To answer these questions, we need to measure how much the domain of different classes overlap. In this subsection, we introduce such measure, based on graph similarity. Given two classes $A$ and $B$, we build a cosine similarity graph $\mathcal{G} = \langle \mathcal{V}, \mathcal{E}, \mathbf{W}\rangle$ of the examples in the latent space. (see Definition~\ref{def:cosine_simi_graph}). The set of vertices gather all samples from both classes: $\mathcal{V}=\X_A\cup\X_B$. After removing self-loops, we only keep the $k|\mathcal{V}|$ heavier edges among the possible ones, i.e. we have $|\Edges|=k|\mathcal{V}|$.

The densely connected graph corresponds to $k=|\mathcal{V}|$ whereas $k=0$ leads to isolated vertices. For an appropriate choice of $k$ somewhere between those two extremes, the resulting graph is sparse and naturally organized into communities with high intra-connectivity. Here, we chose $k=20$.

By applying \emph{Louvain's communities detection algorithm}\cite{blondel2008fast} we detect the cluster $C_{i}\in\C$ associated with the example $i\in\mathcal{V}$. If the classes $A$ and $B$ are well separated, we expect those communities to contain elements of a single class. Conversely, if the domains of classes overlap, those communities will contain different labels. In few-shot setting, due to the lack of examples and the high dimensional nature of data, most algorithms (supervised or unsupervised) rely on the hypothesis that examples in the same neighborhood must share the same label. Hence, we make the hypothesis that a classifier can predict the true label $y\in\{A,B\}$ of the example $i$ knowing its community $C_{i}$. In this case, the \emph{cross entropy} $CE$ of any classifier $\novelClassifier$ against the true distribution $\Prob_{\Y|C}$ is lower bounded by the \emph{conditional entropy}:
\begin{equation}
\begin{split}
CE(\Prob_{\Y|\C}, \novelClassifier)&\geq H(Y|C)\\
                      &=\E_{C\sim\C} \left(H(p_A(C)\right)\\
                      &=\E_{i\sim\mathcal{V}} \left(H(p_A(C_{i}))\right)\,,
\end{split}
\end{equation}
where $Y$ (resp. $C$) is the random variable of labels (resp. communities) under uniform sampling of vertices in $i\in\mathcal{V}$. This lower bound can be computed empirically from the data using the true labels. The frequency $p_A(C_{i})$ of labels $A$ in communities $C_{i}$ is used to compute the binary entropy:
\begin{equation}
\begin{split}
H(p_A(C_i))=&-p_A(C_{i})\log_2{(p_A(C_{i}))}\\
            &-(1-p_A(C_{i}))\log_2{(1-p_A(C_{i}))}\,,
\end{split}
\end{equation}
which reaches its maximum when both classes are equally present in the community, and a minimum when the community only contains one class. Its weighted average over communities is the lower bound. The stochasticity of Louvain's algorithm is counter-balanced by an average over $5$ runs:
\begin{equation}
    S(\X_A,\X_B)=\frac{1}{5}\sum_{r=1}^5H_r(Y|C)\,.
\end{equation}
  
\subsubsection{Results}

The scores $S(\X_A,\X_B)$ are computed for each pair of classes $\{A,B\}$ among the $20$ classes of mini-ImageNet. The features of the images are extracted with \textbf{densenet-m}. The higher the score, the greater the overlap between classes. By keeping the most significant scores, we build the graph of Fig.~\ref{fig:similarity}~(a).

We notice that the mammals are clustered, reflecting an important overlapping between the corresponding classes. Without prior knowledge on the concept of mammal, the backbone suffers from this confusion. They are all connected to each other, meaning they are all sharing some features: it is an example of \emph{class confusion}. The confusion between \textit{Lion} and \textit{African hunting dog} may involve the desert background.  
  
The \textit{Vase} class is an example of \emph{multiple class representatives}: the domain of this class overlaps some other such as \textit{Mixing bowl}, \textit{Cuirass} or \textit{Hourglass}. But there is no transitivity: \textit{Mixing bowl} and \textit{Cuirass} are not very similar to each other. The features of \textit{Vase} are split among different other novel class. Such phenomenon can also be observed with \textit{Nematode}, \textit{Ant} and \textit{King crab}.  
  
\begin{figure}[t]
    \centering
    \includegraphics[width=7cm,height=29cm,keepaspectratio]{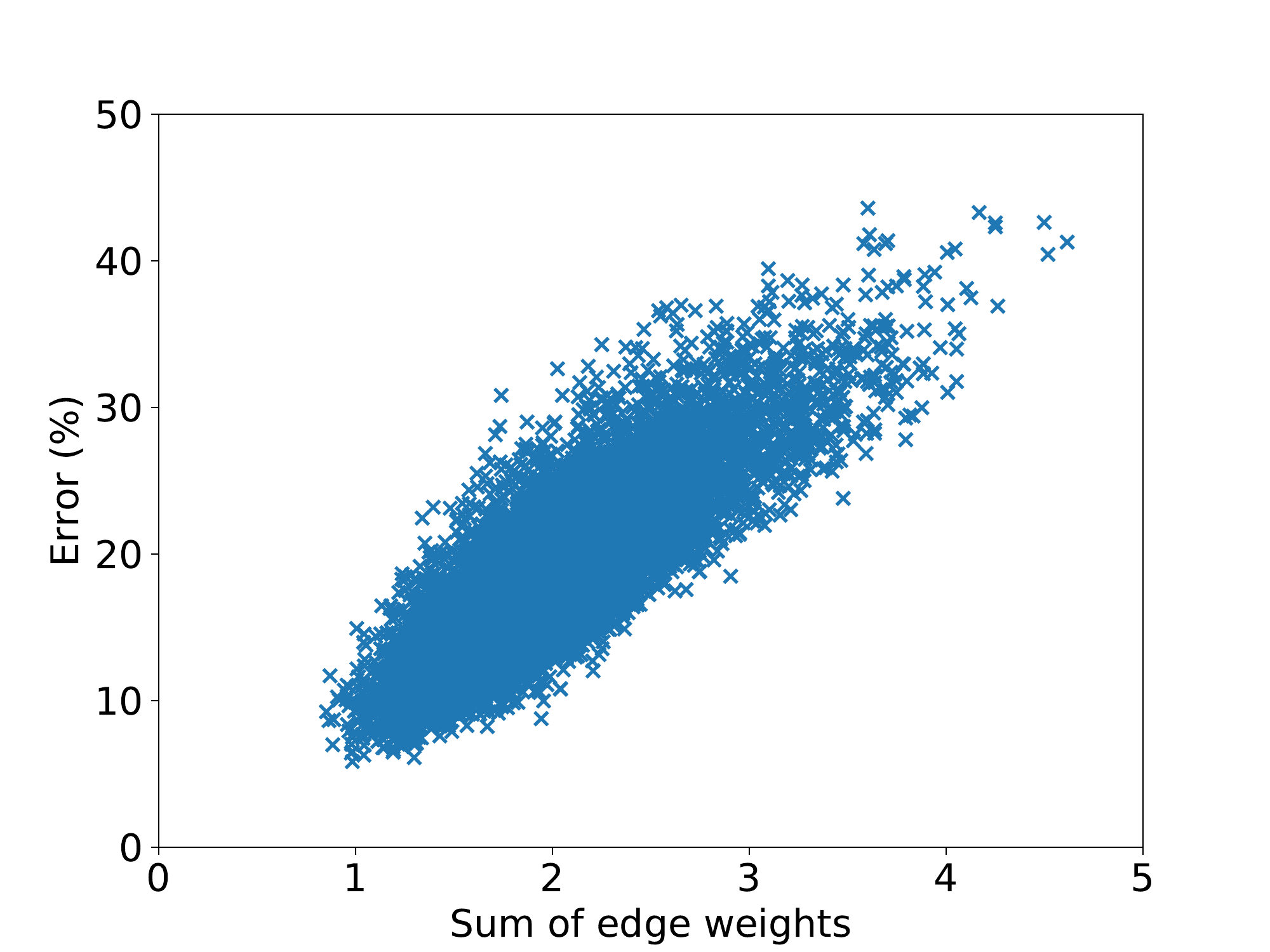}
    \caption{Correlation between a measure of domain overlapping between classes and the error of a LR trained to distinguish the same classes. We generate $10,000$ $5$-way $5$-shot tasks. Each one is represented by a cross. The backbone used is \textbf{densenet-m}, and the classifier a logistic regression. The Pearson correlation coefficient is $0.84$.}
    \label{fig:corrdensenetlogistic}
\end{figure}

Moreover, the score $S(\X_A,\X_B)$ is positively correlated with the average error of a logistic regression. To show it, we uniformly sample $5$ classes and compare the sum of $S(\X_A,\X_B)$ over those five classes with the generalization error in a supervised setting over $5$-way $5$-shot tasks. The obtained points are depicted in Fig.~\ref{fig:corrdensenetlogistic}. The Pearson correlation coefficient is $0.84$, estimated with $10,000$ runs. We choose the $595$ remaining examples for test set, since it gives a more reliable estimate of the performance of the classifier. 

\subsubsection{Influence of the backbone}

A reasonable question is whether the obtained results are specific to the considered backbone. This is why we proceed to the same evaluation with \textbf{wideresnet} and we obtain the graph in Fig.~\ref{fig:similarity} (b).

The two graphs are very similar. On the complete graph (without cutting lighter edges), the correlation edgewise is $0.93$. This high correlation encourages us to think that hard tasks are mostly independent of the backbone and are strongly influenced by the base classes used for backbone training. We wonder if class confusion in novel classes is induced by class confusion with base classes. Said otherwise: \emph{are novel classes confused with base classes sharing similar features?}  
  
\subsubsection{Novel classes are confused with base classes} \label{sec:basenovelconfusion}

To answer this question, we build similarity graphs like the ones of the previous section, involving both base classes and novel classes for \textbf{densenet-m}. We enforce the graph to be bipartite, and we only keep the heavier edges to increase readability (see Fig.~\ref{fig:basenovelconfusion}). We observe mainly three phenomena.  
  
\paragraph{Perfect matching between novel and base class} It is the case for the pairs \textit{bookshop} and \textit{tobacco shop}, and for \textit{ant} and \textit{harvestman}. In this situation, the features of the novel class are entirely determined by the features of \textit{one} base class, which are not split among other novel classes. The two novel classes \textit{bookshop} and \textit{ant} are not part of a cluster nor a hub in Fig.~\ref{fig:similarity} (a). Hence, it seems to be the most desirable case for few-shot learning.  
  
\paragraph{Split of base classes} In this situation, a base class provides features that are shared by many novel classes. It is the case for \textit{saluki} (a specie of dog) which, according to intuition, shares features with two other species of dog (\textit{golden retriever}, \textit{malamute}). Interestingly, the dogs are organized into a hub in Fig.~\ref{fig:similarity} (a). \textit{Vase}, \textit{hourglass} and \textit{cuirass} are also organized into a hub, and they appear to share features with \textit{cocktail shakers}. Hence, the split of base classes seems to be an important cause of the apparition of hubs in Fig.~\ref{fig:similarity} (a).  
  
\paragraph{Split of novel classes} In this situation, the domain of a novel class overlaps the domain of multiple base classes. The most notorious case is the \textit{electric guitar}, which is split among five base classes, including \textit{stage} (which may contain image of guitar), \textit{oboe} (another music instrument), \textit{reel} (many of which are from fishing rods), \textit{holsters} or \textit{clogs}. It is also the case of \textit{mixing bowls}, confused with \textit{frying pans} and \textit{woks}, or \textit{hourglass} with \textit{cocktail shakers} and \textit{chimes}. It seems that the novel classes that are split are involved in more conflicts in Fig.~\ref{fig:similarity} (a).  

\subsection{Conclusion}

We pointed out the problem of the confusion of classes in the feature space when using a transfer-based few-shot solution. The difficulty arises from the overlapping of domains of novel classes that make few-shot approaches inefficient. We developed a measure of domain overlapping (Fig.~\ref{fig:similarity}), and we established its correlation with the accuracy of naive LR (Fig.~\ref{fig:corrdensenetlogistic}). The domain overlapping in latent space depends more on the base classes than on the backbone and its training procedure. It is explained by the fact that novel classes share features with base classes (Fig.~\ref{fig:basenovelconfusion}). 
  
This confusion leads the classifier to overfit on the few available labeled samples. As all labeled samples are used during the training process, the generalization performance of the classifier cannot be checked on another set of samples. The challenge of the following sections consists in predicting the generalization ability of a classifier without having access to more labeled data samples.

\section{Methods} \label{sec:methods}

\begin{table*}[t]
\centering
\caption[]{Table presenting several settings encountered while learning with few labeled samples, and summarizing considered solutions. Solutions are metrics designed to quantify how well a trained model generalizes to unseen data.}
\begin{tabular}{|c|c|c|c|c|}
\cline{3-5}
\multicolumn{2}{c|}{\multirow{2}*{\scalebox{0.7}}} & \multicolumn{3}{c|}{\scalebox{0.8}{\bfseries PROBLEMS}} \\ \cline{3-5}
\multicolumn{2}{c|}{} & \textbf{\textit{Supervised}} &   \textbf{\textit{Semi-supervised}} &  \textbf{\textit{Unsupervised}}  \\
\multicolumn{2}{c|}{} & \textit{$N$-way $K$-shot} &   \textit{$N$-way $K$-shot $Q$-query$^{*}$} & \textit{$N$-way $Q$-query$^{*}$}  \\
\cline{1-5}
\multirow{8}*{\rotatebox{90}{\scalebox{0.9}{\bfseries SOLUTIONS}}}
& \multicolumn{4}{|c|}{\textbf{\textit{Using available labels and features of data samples}}} \\ \cline{2-5}
& \textit{Training loss of a Logistic Regression} & \checkmark & \checkmark & $\times$\\ \cline{2-5}
& \textit{Relative distances between labeled samples} & \checkmark & \checkmark & $\times$ \\ \cline{2-5}
& \textit{Confidence in the output of the classifier} & $\times$ & \checkmark & $\times$ \\
\cline{2-5}
&\multicolumn{4}{|c|}{\textbf{\textit{Using only data relationships}}} \\
\cline{2-5}
&\textit{Eigenvalues of a graph Laplacian} & \checkmark & \checkmark & \checkmark \\
\cline{2-5}
&\textit{Davies-Bouldin score after $N$-means} & \checkmark & \checkmark & \checkmark \\
\cline{1-5}
\multicolumn{5}{l}{$^{*}$Query samples are accessible during training without their labels.}\\
\end{tabular}
\label{tab1}
\end{table*}

Let us remind the setting of the study. Having access to a feature extractor trained on a rich database, a classifier learns to distinguish between new classes using few labeled samples. We aim at quantifying how well the trained classifier generalizes to unseen data.

When performing benchmarks, authors usually have access to more labeled data through a test set. However, in practical applications, when learning from few labeled samples, it is not simple to generate such a set. In this section, we propose several metrics that do not require a test or a validation set, and aiming at quantifying how well a model generalizes to unseen data.
A recap of studied metrics is given in Table~\ref{tab1}.

In the following subsections, we introduce measures for each considered setting that we prove to be correlated with the generalization performance in the next section.

\subsection{Metrics defined on supervised inputs}
Recall that the classifier we use in the experiments is a Logistic Regression (LR). We propose two metrics to estimate how well the trained LR generalizes to unseen data. The first one consists in using the obtained LR training loss at the end of the training process. The second one is based on the relative distances between labeled data samples, and is therefore agnostic of the choice of the LR as classifier. 

\paragraph*{LR training loss} The LR is trained to minimize the cross-entropy loss, see Definition~\ref{def:LRLoss}. During training, this loss is supposed to converge to zero. Assuming that harder the task is, slower the convergence is, the value of the loss at the end of the training should give some insights about the difficulty of the task. Thus, we decide to explore its ability to estimate how easy the generalization to new data samples will be.
\vspace{0.25cm}
\begin{definition}[LR training loss] \label{def:LRLoss}
Given $y_{ic}$ the number (0 or 1) indicating if the label of the data sample $i$ is $c$ and $\mathbf{p}_{ic}$ the output of the LR indicating the probability of $i$ being labeled $c$, the loss is defined as:
\begin{equation}
    \text{LR training loss} = \frac{-1}{NK}\sum_{i=1}^{NK}\sum_{c=1}^{N}{y_{ic}\log\mathbf{p}_{ic}} \,.\label{eq:1}
\end{equation}
\end{definition}
\vspace{0.25cm}

\paragraph*{Similarity metric} Recent works~\cite{hu2020exploiting, wang2019simpleshot} have shown that state-of-the-art performances can be achieved with a decision process that ultimately consists in comparing distances to a centroid defined for each class. Thus, a natural indicator of how easy such a clustering can be performed is to compare the intra-class similarity to the inter-class similarity. In our case, we choose the cosine similarity as the feature vectors are projected onto the unit sphere. We detail the metric we propose by first introducing the notions of intra-class and inter-classes similarities respectively in Definition~\ref{def:intra} and~\ref{def:inter}. Then the proposed metric is written in Definition~\ref{def:similarity}.

\vspace{0.25cm}
\begin{definition}[Intra-class similarity] \label{def:intra}
The cosine similarity within a class $c$ is: \begin{equation}
\text{intra}(c) = \frac{2}{K(K-1)} \sum_{\substack{\phantom{j}i\phantom{j}\\ y_i=c}}\sum_{\substack{j\ne i\\ y_j=c}}{ \text{cos}(\mathbf{f}_i, \mathbf{f}_j)}\,.
\end{equation}
If a class $c$ only contains one shot, we set $\text{intra}(c) = 1$.
\end{definition}
\vspace{0.25cm}

\begin{definition}[Inter-classes similarity] \label{def:inter}
The cosine similarity through classes $c$ and $\tilde{c}$ is:
\begin{equation}\text{inter}(c, \tilde{c}) = \frac{1}{K^2} \sum_{\substack{i\phantom{j}\\y_i=c\phantom{\tilde{c}}}}\sum_{\substack{j\\y_j=\tilde{c}}}{\text{cos}(\mathbf{f}_i, \mathbf{f}_j)}\,.
\end{equation}
\end{definition}
\vspace{0.25cm}

\begin{definition}[Similarity metric] \label{def:similarity}
The proposed similarity metric is:
\begin{equation}
    \text{similarity} = \frac{1}{N}\sum_{c=1}^{N}{\left(\text{intra}(c) - \max_{c \ne \tilde{c}}(\text{inter}(c, \tilde{c}))\right)}\,.\label{eq:4}
\end{equation}
\end{definition}
\vspace{0.25cm}

\subsection{Metrics defined on unsupervised inputs}
In the unsupervised setting, we use the $N$-means algorithm. Again, the goal is to estimate the quality of the obtained clustering. To this end, we consider two metrics. The first one is based on a relative similarity measure between clusters. The other one is an indirect measure of component connectivity in a graph whose vertices are the considered samples and edges represent the similarity between those samples.

\paragraph*{Davies-Bouldin score after $N$-means} Assuming the data samples within classes are similar enough, we expect each learned cluster to represent a class. A measure of relative similarity within clusters and between clusters, such as the classical Davies-Bouldin (DB) score~\cite{davies1979cluster}, gives an insight about the difficulty of the clustering. Consequently, it may measure how easy it is to generalize to new samples. In Definition~\ref{def:DBScore}, we detail the Davies-Bouldin (DB) score. Lower is the score, better is the clustering. It varies between $0$ and $+\infty$.

\vspace{0.25cm}
\begin{definition}[Davies-Bouldin score]\label{def:DBScore} Denote the centroid of a cluster $C_c$ $\bm{\mu}_c$, such that $\bm{\mu}_c = \frac{1}{|C_c|}\sum_{i\in C_c}{\mathbf{f}_i}$. The average distance between samples and the centroid of their cluster is: 
\begin{equation}
\bm{\delta}_c = \frac{1}{|C_c|}\sum_{i\in C_c}{\norm{\mathbf{f}_i - \bm{\mu}_c}_2}\,.
\end{equation}
Then, the DB score is given as:
\begin{equation}
    \text{DB score} = \frac{1}{N}
    \sum_{c=1}^{N}{
    \max_{c\ne \tilde{c}}
    \left( 
    \frac{\bm{\delta}_c + \bm{\delta}_{\tilde{c}}}
    {\norm{\bm{\mu}_c - \bm{\mu}_{\tilde{c}}}_2} \right)
    }\,.\label{eq:3}
\end{equation}
\end{definition}
\vspace{0.25cm}

\paragraph*{Laplacian eigenvalues} Consider a graph where each vertex represents a data sample and edges are weighted according to a similarity between these samples. In a perfect case where samples from distinct classes are very dissimilar, it is expected that this graph yields at least as many connected components as the number of classes in the considered problem. A measure of the fact a graph contains at least $N$ connected components is given by the amplitude of the $N$-th lower eigenvalue of its Laplacian~\cite{shuman2013emerging}. See Definition~\ref{def:egv}.

\vspace{0.25cm}
\begin{definition}[$N$-th eigenvalue]\label{def:egv}
We consider the graph $\mathcal{G}~=~\langle\mathcal{V}, \mathcal{E},\mathbf{W}\rangle$ where $\mathcal{V}$ is the set of data samples. The adjacency matrix $\mathbf{W}$ is obtained by first considering the cosine similarity between these samples, removing self-loops, and keeping only the $k$-th largest values on each line/column. See Definition~\ref{def:cosine_simi_graph} for more details. The Laplacian of the graph is given by $\mathbf{L} = \mathbf{D} - \mathbf{W}$, where $\mathbf{D}$ is the degree matrix of the graph: $\mathbf{D}$ is a diagonal matrix where $\mathbf{D}_{ii} = \sum_{j=1}^{NQ}{\mathbf{W}_{ij}}$. 
The measure we consider is the amplitude of the $N$-th lower eigenvalue of $\mathbf{L}$.
\end{definition}
\vspace{0.25cm}

\subsection{Metric defined in the semi-supervised setting}

Consider we have access to a LR classifier and unlabeled inputs.
We propose a metric based on the confidence of the LR decision on the unlabeled samples. The confidence can be obtained by looking at the distance between the provided output and a one-hot-bit encoded version of this output.

In more details, for each unlabeled sample, the classifier outputs the probability it belongs to a particular class. As we do not know the label of the sample, we cannot look at the probability the classifier gives to its real class. However, we propose to report the maximal probability the classifier attributes to the classes (see Definition~\ref{def:confidence}). If the maximal probability is far from one, we can interpret it as the classifier is unsure of its output. So, lower the maximal probability is, harder the task could be for the considered sample.

\vspace{0.25cm}
\begin{definition}[LR confidence]\label{def:confidence}
Let $\mathbf{p}_{ic}$ denote the probability that the data sample $i$ is labeled $c$.
\begin{equation}
    \text{LR confidence} = \frac{-1}{NQ}\sum_{i=1}^{NQ}{\log \max_c(\mathbf{p}_{ic})}\,. \label{eq:5}
\end{equation}
\end{definition}
\vspace{0.25cm}

In the next section, we empirically measure how correlated are the proposed measures with generalization performance. We also propose to predict the latter from the former.

\section{Experiments} \label{sec:experiments}
In this section, we evaluate the interest of metrics proposed in Section~\ref{sec:methods}.
We perform experiments on data samples coming from mini-ImageNet with features extracted using \textbf{wideresnet} and on data samples from tiered-ImageNet with features extracted using \textbf{densenet-t}. In the unsupervised and semi-supervised settings, let us recall that the extracted features are diffused through a similarity graph before being used. More details on the datasets and the backbones are in Section~\ref{sec:background}.

As we observe that the relations between the metrics and the accuracies on the test sets are rather linear (see Fig.~\ref{fig:exp-super-2} and Fig.~\ref{fig:exp-semi-1}), we report in subsections~\ref{subsec:supervised}, \ref{subsec:unsupervised}, and \ref{subsec:semi} the absolute values of the Pearson correlation coefficients between metrics and accuracies on various settings. In subsection~\ref{subsec:predict}, we try to predict directly the accuracy of a task on a set of data samples whose labels were unknown during the training process. 
Subsection~\ref{subsec:additional} explores the impact of some parameters used in our experiments. Finally, in a last subsection, we investigate whether the LR confidence metric can be used to increase the accuracy of few-shot classification by selecting the samples to annotate.

\subsection{Supervised setting}
\label{subsec:supervised}

\begin{figure}[t]
\centerline{
\begin{adjustbox}{width=0.5\textwidth}
\input{Figures/supervised/fig1}
\end{adjustbox}}
\caption{Supervised setting. Study of linear correlations between the metrics and the accuracy of a LR computed on a test set. In (a) and (b), the data come from mini-ImageNet. Their features are extracted with \textbf{wideresnet}. In (c) and (d), the data come from tiered-ImageNet. Their features are extracted with \textbf{densenet-t}. See Section~\ref{sec:background} for details. By default, $5$-way $5$-shot tasks are generated. In (a) and (c), the number of shots varies. In (b) and (d), the number of classes varies. Each point is obtained over $10,000$ random tasks.}
\label{fig:exp-super-1}
\end{figure}
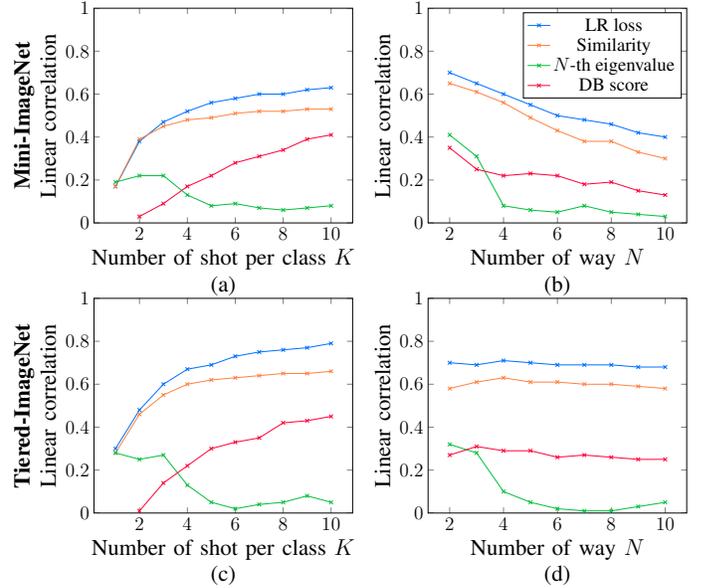

We begin with the supervised setting. We propose to study the linear correlation between the metrics defined on supervised inputs computed on a $N$-way $K$-shot task and the accuracy of the LR on a test set. The test set gathers $50$ unseen data samples per class. We also look at the correlations obtained with the metrics defined on unsupervised inputs, considering all training samples are unlabeled. In Fig.~\ref{fig:exp-super-1}, we perform experiments on mini-ImageNet ((a) and (b)) and on tiered-ImageNet ((c) and (d)). In (a) and (c), we consider $5$-way tasks and depict the evolution of the correlation as a function of the number of shots per class. In (b) and (d), we consider $5$-shot tasks and make the number of classes vary. Note that in $1$-shot tasks, as the DB score is always $0$, we do not report a correlation measure. In Appendix~\ref{app:performances}, the average accuracies of the LR for each type of task are given.

\textbf{Observations:} In all experiments, we observe that the metrics adapted to the supervised setting (i.e. LR loss and similarity) perform better than the metrics designed for an unsupervised setting. The best correlation is always obtained with the LR loss. About the LR loss and the similarity, the more shots there are, the better the correlations are. The more ways there are, the worse the correlations are on mini-ImageNet. On tiered-ImageNet, they do not vary.

\textbf{Discussion:} The metrics designed for a supervised setting overcome the ones designed for an unsupervised setting. Indeed, the supervised metrics exploit an additional information given by the labels. Moreover, the unsupervised metrics are here used on very few data samples. In the next subsections, we show the results they can obtain using more unlabeled samples. The number of shots and ways influence the complexity of the tasks. Higher the number of ways is or lower the number of shots is, harder the task becomes. The linear correlation seems to be lower when the tasks are harder.

\begin{figure}[t]
\centerline{
\begin{adjustbox}{width=0.5\textwidth}
\input{Figures/supervised/fig2}
\end{adjustbox}}
\caption{Supervised setting. Each point represents a task. We plot the accuracy of a LR in function of the loss of the LR on the training samples. In (a), we consider $10,000$ random $5$-way $5$-shot tasks. In (b), we consider $10,000$ random $5$-way $1$-shot tasks. The data samples come from mini-ImageNet. Their features are extracted with \textbf{wideresnet}.}
\label{fig:exp-super-2}
\end{figure}
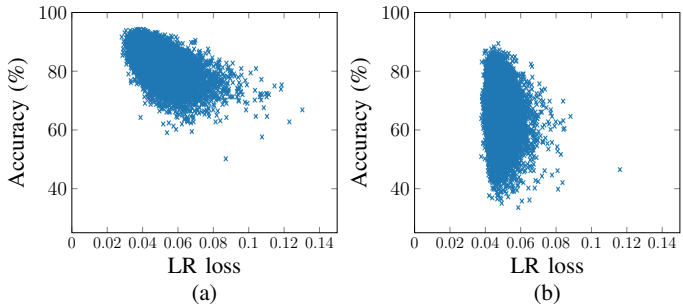

We investigate what happens in Fig.~\ref{fig:exp-super-2}. We generate two plots using data samples from mini-ImageNet. Each point represents a task, with the LR loss on the X-axis and the accuracy on the Y-axis. In (a), $5$-way $5$-shot tasks are considered. In (b), $5$-way $1$-shot tasks. In $5$-way $5$-shot, the relation between both variables is rather linear. Without surprise, in $5$-way $1$-shot, the LR loss is less representative of the accuracy. With only one sample per class, it is very hard to detect the problematic tasks.

\subsection{Unsupervised setting}
\label{subsec:unsupervised}

\begin{figure}[ht]
\centerline{
\begin{adjustbox}{width=0.5\textwidth}
\input{Figures/unsupervised/fig1}
\end{adjustbox}}
\caption{Unsupervised setting. Study of linear correlations between the metrics and the ARI of a $N$-means algorithm. The ARI is computed on the $NQ$ unlabeled samples available during training. In (a) and (b), the data come from mini-ImageNet. Their features are extracted with \textbf{wideresnet}. In (c) and (d), the data come from tiered-ImageNet. Their features are extracted with \textbf{densenet-t}. All features are diffused through a similarity graph. See Section~\ref{sec:background} for details. By default, $5$-way $35$-query tasks are generated. In (a) and (c), the number of queries varies. In (b) and (d), the number of classes varies. Each point is obtained over $10,000$ random tasks.}
\label{fig:exp-unsup-1}
\end{figure}
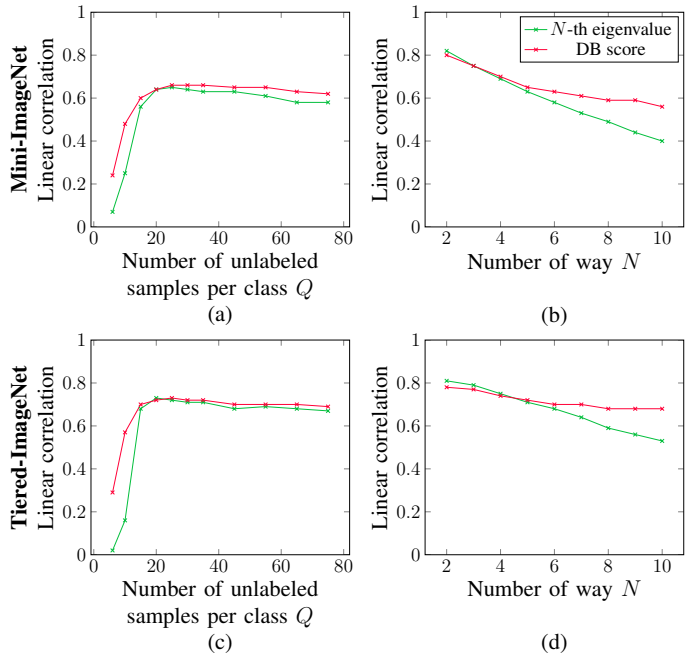

In this setting, we consider $N$-way $Q$-query tasks. We study the linear correlations between the metrics and the adjusted rand index (ARI) of a $N$-means algorithm. The ARI is computed on the unlabeled samples available when training the $N$-means algorithm. In Fig.~\ref{fig:exp-unsup-1}, we perform experiments on mini-ImageNet ((a) and (b)) and on tiered-ImageNet ((c) and (d)). In (a) and (c), we consider $5$-way tasks. The number of queries varies. In (b) and (d), we consider $35$-query tasks. The number of classes $N$ varies. In Appendix~\ref{app:performances}, the average ARI of the $N$-means for each type of task are given.

\textbf{Observations:} In all experiments, we observe that the DB-score is the best. The more queries there are, the better the correlation is, although there is a threshold after $20$ queries per class. The correlation with the $N$-th eigenvalue slightly decrease but this is an artifact due to the construction of the graph. We propose, in Appendix~\ref{app:neighbors}, an experiment showing the influence of the number of nearest neighbors kept in the graph used to compute the eigenvalue. We also observe that the more classes there are, the lower the correlation becomes.

\textbf{Discussion:} As the DB-score is more aligned to the $N$-means algorithm, its correlation with the ARI is higher. We notice that $20$ samples per class are sufficient to increase the correlation up to $0.64$ on mini-ImageNet and up to $0.72$ on tiered-ImageNet. It seems that using more samples is useless. As the complexity of the tasks increases with the number of ways, our metrics are less representative of the problems.

\begin{figure*}[t]
\centerline{
\begin{adjustbox}{width=\textwidth}
\input{Figures/semi-supervised/fig1}
\end{adjustbox}}
\caption{Semi-supervised setting. Study of linear correlations between the metrics and the accuracy of a LR on the $NQ$ unlabeled samples available during training. In (a), (b), (c) and (d), the data samples come from mini-ImageNet. Their features are extracted with \textbf{wideresnet} and diffused through a similarity graph. In (e), (f), (g) and (h), the samples come from tiered-ImageNet. Their features are extracted with \textbf{densenet-t} and diffused though a similarity graph. See Section~\ref{sec:background} for details. By default, $5$-way $5$-shot $30$-query tasks are generated. In (a) and (e), the number of queries varies. In (b) and (f), the number of shots varies. In (c) and (g), the number of classes varies. Each point is obtained over $10,000$ random tasks. In (d) and (h), each point represents a task. We plot the accuracy of the LR in function of the LR confidence. In (d), $5$-way $5$-shot $30$-query tasks are generated from mini-ImageNet. In (h), $5$-way $5$-shot $30$-query tasks are generated from tiered-ImageNet.}
\label{fig:exp-semi-1}
\end{figure*}
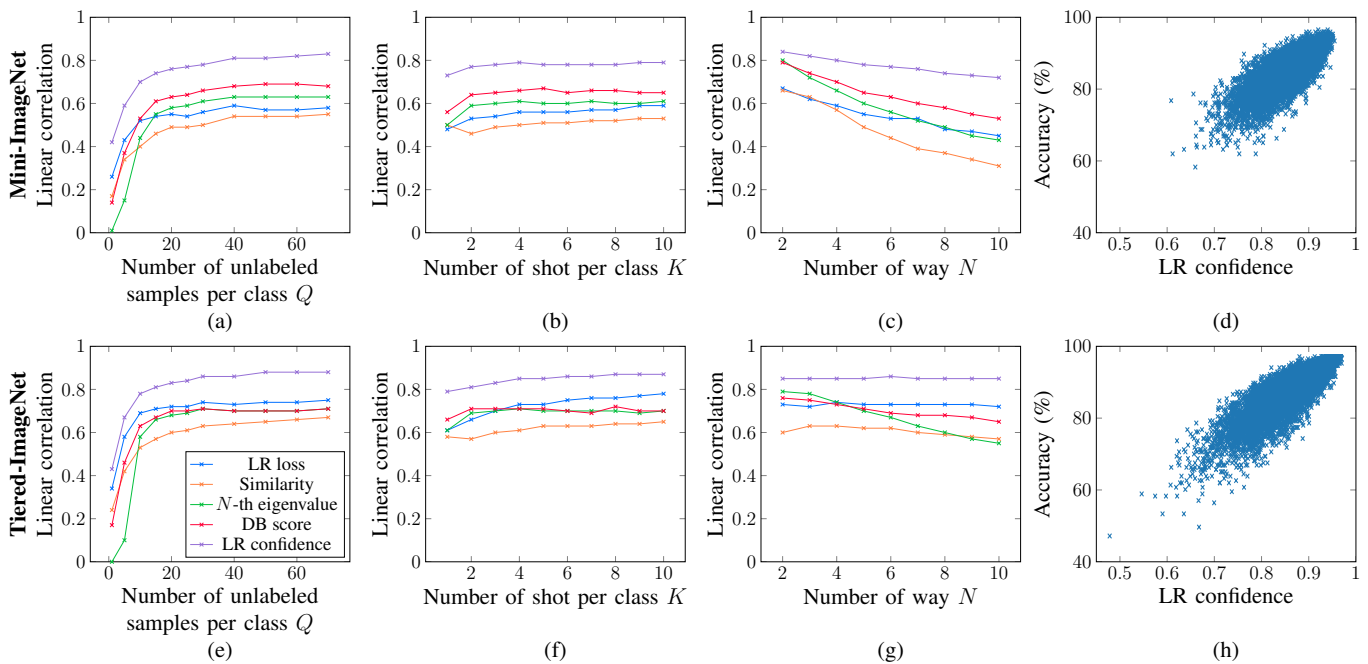

In this section, we do not have access to the labels of the samples during the training but, we make sure that each class contains as many samples. In subsection~\ref{subsec:additional}, we explore the impact of an unbalanced distribution. We also propose an additional experiment to see if the $N$-th eigenvalue is really the one giving the best correlation among all other eigenvalues.

In the next subsection, we wonder to what extent the linear correlations can be increased in a semi-supervised setting.

\subsection{Semi-supervised setting}
\label{subsec:semi}

We consider $N$-way $K$-shot $Q$-query tasks. The query samples are available without their labels during training. We study the linear correlation between the LR confidence and the accuracy of the LR on the query samples. In practice, on $NQ$ samples. We also look at the correlations obtained with the metrics defined on supervised and unsupervised inputs. In the unsupervised case, we consider all training samples as unlabeled samples. In Fig.~\ref{fig:exp-semi-1}, we perform experiments on mini-ImageNet ((a), (b), (c) and (d)) and on tiered-ImageNet ((e), (f), (g) and (h)). In (a) and (e), we consider $5$-way $5$-shot tasks. The number of queries varies. In (b) and (f), we consider $5$-way $30$-query tasks. The number of shots varies. In (c) and (g), we consider $5$-shot $30$-query tasks. The number of classes varies.
In (d) and (h), there are two scatter plots. Each point represents a task, with the LR confidence on the X-axis and the accuracy on the Y-axis. In both cases, $5$-way $5$-shot $30$-query tasks are considered. In Appendix~\ref{app:performances}, the average accuracies of the LR for each type of task are given.

\textbf{Observations:} We observe that the LR confidence, a metric adapted to the semi-supervised setting, outperforms the supervised and the unsupervised metrics. 
The supervised metrics seem to depend on the number of unlabeled samples per class $Q$. This is, in part, an artifact due to the fact that the accuracies are computed on the number of queries $NQ$. If $Q$ is small, the range of possible accuracies is reduced, so the computed correlation is affected. The more queries there are, the better the correlation is, with a threshold around $15$ additional unlabeled samples per class. The correlation depends less on the number of shots and classes. We observe that the correlation between the supervised metrics and the accuracy are higher than in the supervised setting for the small $K$. This is due to the fact that the features are previously diffused on a similarity graph.

\textbf{Discussion:} The LR confidence uses more information than the unsupervised metrics (labels) and the supervised metrics (more data). As in all metrics, a small number of shots does not reduce the linear correlation, we assume that this is due to the diffusion of the features before the training of the LR.

As in the unsupervised setting, we do not have access to the labels of the samples during the training but, we use as many samples per class. In subsection~\ref{subsec:additional}, we explore the impact of an unbalanced distribution. We also propose an additional experiment on the relevance of different eigenvalues. In subsection~\ref{subsec:annotate}, in order to see if we can use the LR confidence to increase the accuracy of the LR on some tasks, we propose to label the data samples with the lowest confidences.

\subsection{Predicting task accuracy}
\label{subsec:predict}

\begin{figure}[t]
\hspace{-0.5cm}
    \centerline{
    \begin{adjustbox}{width=0.2\textwidth}
    \input{Figures/prediction/conf-notations}
    \end{adjustbox}}
    \caption{Notations}
    \label{fig:exp-roc-1}
\end{figure}

We want to study the relevance of some metrics as proxies to the generalization performance on a set of unseen labeled data samples. In the previous subsections, we look at the linear correlations between the metrics and the performances on a test set. They measure the joint variation between both variables, but not the ability of the metrics to predict the performances. To go further in this section, we evaluate to what extent the metrics can predict the accuracy of a classifier.

We consider three settings: supervised, unsupervised and semi-supervised. In the supervised setting, we try to predict the accuracy of a LR. In the unsupervised setting and semi-supervised setting, we try to predict the accuracy of a LR when the features have been previously diffused over a similarity graph. To keep the experiments as interpretable as possible, we propose a binary classification problem. The tasks having an accuracy below $80\%$ are \emph{hard} and the ones above $80\%$ are \emph{easy}, $80\%$ being an arbitrary choice.

In each setting, we wonder which threshold value of metrics enables to distinguish the best between hard and easy tasks. We plot a ROC curve. The x-coordinate is $1\text{ -- specificity}$. It indicates the proportion of tasks that have been classified as \emph{hard} tasks among the \emph{easy} tasks. The y-coordinate is the sensibility. It indicates the proportion of tasks that have been classified as \emph{hard} tasks among the \emph{hard} tasks. Knowing these two variables, we could choose an adequate threshold value. Following the notations in Fig.~\ref{fig:exp-roc-1}, we have:
\begin{equation}
    1\text{ -- specificity} =
    \frac{\text{FP}}{\text{TN}+\text{FP}}\ ,
\end{equation}
\begin{equation}
    \text{sensibility} = \frac{\text{TP}}{\text{TP}+\text{FN}}\ .
\end{equation}

In practice, we divide the mini-ImageNet dataset into two sets. Both containing $10$ classes. On both sets, $10,000$ $5$-way $5$-shot ($30$-query) tasks are randomly generated. In the unsupervised case, we consider all training samples as unlabeled samples. In Fig.~\ref{fig:exp-roc-1}, we plot the ROC curve using the first set of tasks. After choosing the threshold value, we report a confusion matrix on the second set. We normalize each row of the confusion matrix, so that its first row indicates the percentage of tasks predicted hard or easy among the hard tasks and its second row indicates the percentage of tasks predicted hard or easy among the easy tasks.

\begin{figure}[t]
    \centering
    
    \begin{subfigure}[t]{0.25\textwidth}
    \centerline{
    \begin{adjustbox}{width=\textwidth}
    \input{Figures/prediction/pred-super}
    \end{adjustbox}}
    \caption{Supervised setting.}
    \end{subfigure}\quad
    \begin{subfigure}[t]{0.20\textwidth}
    \centerline{
    \begin{adjustbox}{width=\textwidth}
    \input{Figures/prediction/conf-super}
    \end{adjustbox}}
    \end{subfigure}
    \\
    \begin{subfigure}[t]{0.25\textwidth}
    \centerline{
    \begin{adjustbox}{width=\textwidth}
    \input{Figures/prediction/pred-unsup}
    \end{adjustbox}}
    \caption{Unsupervised setting.}
    \end{subfigure}\quad
    \begin{subfigure}[t]{0.20\textwidth}
    \centerline{
    \begin{adjustbox}{width=\textwidth}
    \input{Figures/prediction/conf-unsup}
    \end{adjustbox}}
    \end{subfigure}
    \\
    \begin{subfigure}[t]{0.25\textwidth}
    \centerline{
    \begin{adjustbox}{width=\textwidth}
    \input{Figures/prediction/pred-semi}
    \end{adjustbox}}
    \caption{Semi-supervised setting.}
    \end{subfigure}\quad
    \begin{subfigure}[t]{0.20\textwidth}
    \centerline{
    \begin{adjustbox}{width=\textwidth}
    \input{Figures/prediction/conf-semi}
    \end{adjustbox}}
    \end{subfigure}
    \caption{Task prediction. The ROC curves are computed over a subset of mini-ImageNet containing $10$ classes. The tables are computed on the remaining $10$ classes, applying the threshold value denoted by a red point on the curves. Features are extracted with \textbf{wideresnet}. In both cases, $10,000$ $5$-way $5$-shot ($30$-query) are randomly generated. In (a), the variable is the LR loss, in (b), the DB-score, and in (c), the LR confidence.}
    \label{fig:exp-roc-2}
\end{figure}
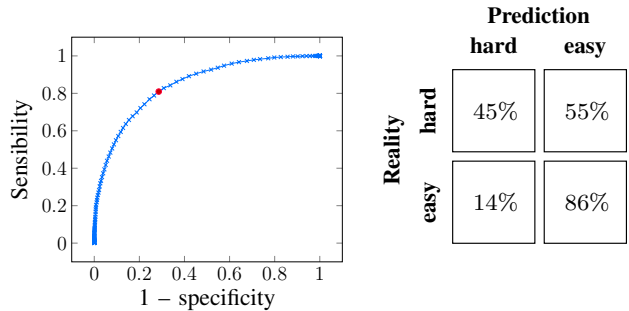
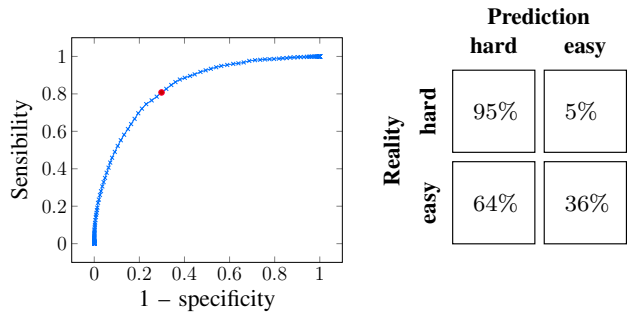
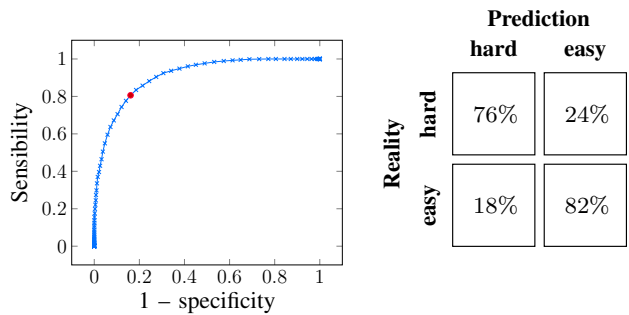

\textbf{Supervised setting:} In Fig.~\ref{fig:exp-roc-2} (a), the ROC curve is built by varying a threshold value over the LR loss. When selecting a threshold value at $(0.29, 0.81)$, we obtain a confusion matrix on the second set where $1\text{ -- specificity}$ becomes $0.14$ and sensibility becomes $0.45$. As both variables are lower, the chosen threshold does not apply to the second set.

\textbf{Unsupervised setting:} In Fig.~\ref{fig:exp-roc-2} (b), the ROC curve is built by varying a threshold value over the DB-score. When selecting a threshold value at $(0.30, 0.81)$, we show on the confusion matrix that $1\text{ -- specificity}$ becomes $0.64$ and sensibility becomes $0.95$. Here, both variables are higher. Once again, the chosen threshold does not apply to the second set.

\textbf{Semi-supervised setting:} In Fig.~\ref{fig:exp-roc-2} (c), the ROC curve is built by varying a threshold value over the LR confidence. We select the threshold value at $(0.16, 0.81)$. In the confusion matrix, $1\text{ -- specificity}$ becomes $0.76$ and sensibility becomes $0.18$. Both variables are similar on the two sets. So, the chosen threshold value generalizes to the second set.

Moreover, we perform a second experiment considering only the LR confidence in a semi-supervised setting. The advantage of the LR confidence over other metrics is that it is easily interpretable. Indeed, it associates with each task the average confidence of the LR on each query sample. We propose to consider the LR confidence value as a predicted accuracy. We perform an experiment on $10,000$ $5$-way $5$-shot tasks. The average error between the real and the predicted accuracies is $2.40\%$. To evaluate this result, we also compute the mean absolute deviation of the real accuracies from their average. It amounts to compare the LR confidence with a naive method always predicting the same accuracy. The mean absolute deviation is $4.14\%$. Thus, the predictions of the LR confidence are better than the naive method.

\subsection{Additional experiments exploring different parameters}
\label{subsec:additional}

\begin{figure}[!ht]
\centerline{
\begin{adjustbox}{width=0.5\textwidth}
\input{Figures/unbalanced}
\end{adjustbox}}
\caption{Influence of the proportion of unlabeled data samples $p$ in a class with respect to the other ones. The features are extracted with \textbf{wideresnet} from mini-ImageNet, and diffused through a similarity graph. In the semi-supervised settings (a-b), we report linear correlations between the metrics and the accuracy of a LR on the unlabeled samples. In the unsupervised settings (c-d), we report linear correlations between the metrics and the ARI of a $N$-means algorithm. In (a-c), $2$-way $5$-shot $50$-query tasks are generated. In (b-d), $5$-way $5$-shot $50$-query. The proportion of samples in the other classes is identical. Each point is obtained over $10,000$ random tasks.}
\label{fig:exp-unbalanced}
\vspace*{0.75cm}
\centerline{
\begin{adjustbox}{width=0.5\textwidth}
\input{Figures/egv}
\end{adjustbox}}
\caption{Analysis of the relevance of eigenvalues with different number of classes. The features are extracted with \textbf{wideresnet} from mini-ImageNet, and diffused through a similarity graph. In the semi-supervised setting (a), $5$-shot $30$-query tasks are generated. In the unsupervised setting (b), $35$-query tasks. We report linear correlations between the $N$-th eigenvalue and the accuracy of the LR (a) / the ARI of the $N$-means algorithm (b). In both settings, we also report the index of the eigenvalue which enables the best correlation. Each point is obtained over $10,000$ random tasks.}
\label{fig:exp-egv}
\end{figure}

In previous experiments, they are as many unlabeled samples per class. We propose to explore what happens when their distribution is unbalanced between classes.

In Fig.~\ref{fig:exp-unbalanced}, first column, we propose an experiment on $2$-way $5$-shot $50$-query tasks, where we vary the proportion of unlabeled data samples in the first class with respect to the second one. In the unsupervised case, we consider that all samples are unlabeled. In both semi-supervised (a) and unsupervised (c) settings, the linear correlations tend to decrease with the imbalance. The supervised metrics (LR loss and similarity) do not take into account the unlabeled samples. However, whereas they are computed on balanced labeled samples, the accuracy is computed on an unbalanced set of samples. This may explain the decrease. The linear correlations with the LR confidence are rather constant. As the LR confidence is directly linked to the query samples and the LR, it is more robust. In the semi-supervised setting, the linear correlations with the DB-score and the $N$-th eigenvalue goes to $0$. The DB-score measures the quality of the $N$ clusters made by a $N$-means on the unlabeled samples. The $N$-th eigenvalue measures to what extent a $15$ nearest neighbors similarity graph computed on the unbalanced data samples is far from having $N$ connected components. When the distribution of the unlabeled samples is unbalanced, these measures on clusters/connected components no longer represent what happens in the LR. In the unsupervised setting, the DB-score stays correlated with the ARI.

In Fig.~\ref{fig:exp-unbalanced}, second column, we propose an experiment on $5$-way $5$-shot $50$-query tasks, where we vary the proportion of unlabeled data samples in a class with respect to the four other classes. The four classes keep the same number of samples. When the number of unlabeled samples in the first class is closed to $0$, the problem amounts to a $4$-way classification problem with balanced samples, so the correlations are not really influenced. When the number of unlabeled samples in the first class goes to $1$, the performances of all metrics, except the LR confidence, decrease. The same reasons as in the $2$-way experiment explain the results.

When we motivated the use of the $N$-th lower eigenvalue as a metric, we assumed that the $N$ classes should correspond to $N$ connected components in a graph whose edges only connect the most similar samples. However, in practice, it is expected that the number of components differ and as such more useful information could be carried by other eigenvalues. In Fig.~\ref{fig:exp-egv}, we report the linear correlation in function of the number of ways where the $N$-way eigenvalue and the best one are plotted. In both semi-supervised and unsupervised settings, the index of the best eigenvalue is lower than the one expected. The performance gap increases with the number of ways.

\subsection{Using per-sample confidence to choose samples to label}
\label{subsec:annotate}
\begin{figure}[t]
    \begin{adjustbox}{width=0.45\textwidth}
    \input{Figures/annotation/annotation}
    \end{adjustbox}
    
    \caption{Using per-sample confidence to label data. In a semi-supervised setting, we label some of the unlabeled samples available during training, either randomly or based on the lowest confidences attributed by the LR. At first, we consider $5,000$ random $5$-way $1$-shot $50$-query tasks. We recompute the accuracies obtained after labeling some samples. The data come from mini-ImageNet. Their features are extracted with \textbf{wideresnet} and diffused through a similarity graph.}
    \label{fig:exp-annotate}
\end{figure}
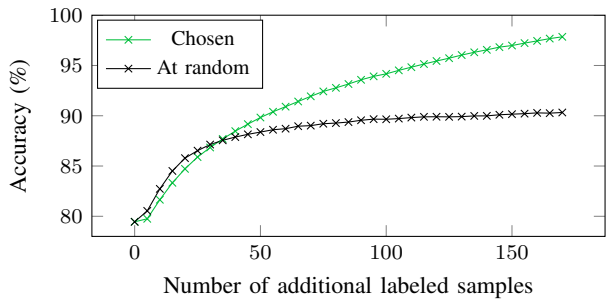

Additionally, we propose a last experiment in a semi-supervised setting. Using the LR confidence, we can attribute to each query sample a confidence value. This confidence can be used to decide which unlabeled samples should be labeled for increasing the accuracy of the classifier. In Fig.~\ref{fig:exp-annotate}, we compare what happens when labeling specific query examples, and when labeling examples at random. We observe that when the number of labeled samples is small, it is better to have a random selection. This is due to the fact the sampling is more balanced between classes than when using the proposed method. Above a certain amount of labeled samples, it becomes clearly more efficient to choose which samples to label. This is not surprising as the chosen elements happen to be the ones with the lowest confidences, meaning that the remaining ones are easy to classify.

\section{Conclusion} \label{sec:conclusion}

In this paper, we introduced the problem of predicting the generalization performance of few-shot solutions, taking into account the fact they do not have access to a large enough validation set. We first identified that the difficulty of the considered task relies mainly on the relation between base classes, used to train the feature extractor, and novel classes. We then introduced several measures that we showed to be correlated to generalization performance in various settings: supervised, unsupervised and semi-supervised. Interestingly, one of these measures can predict quite well the quality of generalization, despite the lack of labeled data.
In future work, we would like to inquire in more details how these findings could help in designing more efficient solutions for the few-shot problem, for example by choosing which samples to label.

\newpage
\begin{appendices}
\section{Performances of models on various tasks}
\label{app:performances}

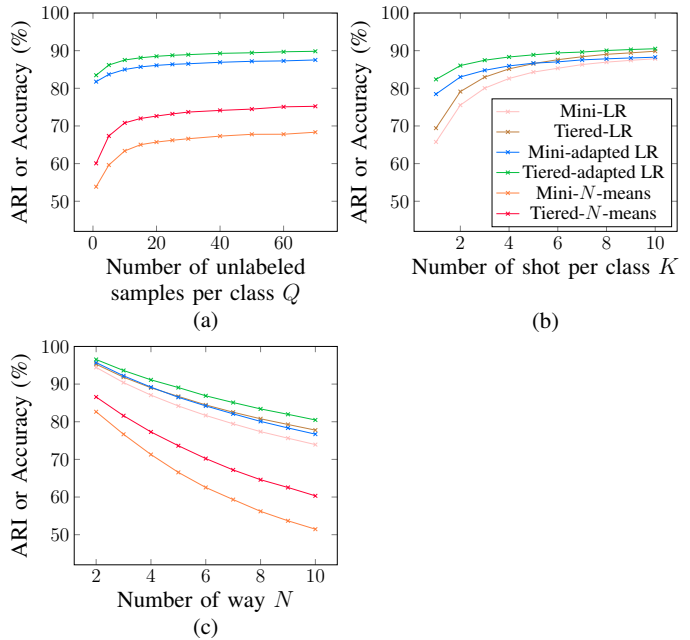
\begin{figure}[ht]
\centerline{
\begin{adjustbox}{width=0.5\textwidth}
\input{Figures/accuracy}
\end{adjustbox}}
\caption{We report the performances of models used in Fig.~\ref{fig:exp-super-1}, \ref{fig:exp-unsup-1} and \ref{fig:exp-semi-1}. By default, $5$-way $5$-shot $30$-query tasks are generated. Mini means that data come from mini-ImageNet. Tiered means that data come from tiered-ImageNet. Mini-LR and Tiered-LR are the accuracies obtained in the supervised setting. Mini-adapted LR and Tiered-adapted LR, the accuracies obtained in the semi-supervised setting. Mini-$N$-means and Tiered-$N$-means are the ARIs obtained in the unsupervised setting. For reasons of scale, we multiply by $100$ the ARIs.}
\label{fig:exp-acc}
\end{figure}

\vspace{-0.5cm}
\section{Influence of the number of nearest neighbors}
\label{app:neighbors}

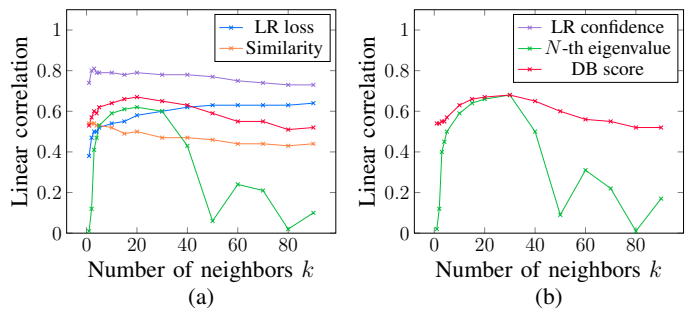
\begin{figure}[ht]
\centerline{
\begin{adjustbox}{width=0.5\textwidth}
\input{Figures/knn}
\end{adjustbox}}
\caption{Analysis of the influence of the number of neighbors $k$ kept in the similarity graphs. Data come from mini-ImageNet. Their features are extracted with \textbf{wideresnet} and diffused through a $k$-nearest neighbors similarity graph. $5$-way $5$-shot $30$-query tasks are generated. In (a), the correlations are computed between the metrics and the accuracy of a LR on the unlabeled samples. In (b), they are computed between the metrics and the ARI of a $N$-means on the unlabeled samples. Each point is obtained over $10,000$ random tasks.}
\label{fig:exp-knn}
\end{figure}
\end{appendices}

\bibliographystyle{plain}
\bibliography{references}

\vspace{12pt}

\end{document}

%% file: Pseudo-codes/Random.tex
\removelatexerror
\begin{algorithm}[H] \label{alg:random}
\caption{Standard deviation of the accuracy}
\begin{algorithmic}
\STATE $perf$ $\leftarrow$ empty list
\FOR{run $=1 \rightarrow 500$}
\STATE Select randomly $5$ classes and $5$ shots within classes.
\STATE Generate a test set with $50$ other samples per class.
\STATE Train a LR to classify the $5$ classes from the features.
\STATE Append the LR accuracy on the test set to $perf$.
\ENDFOR
\RETURN standard\_deviation($perf$)
\end{algorithmic}
\end{algorithm}

%% file: Pseudo-codes/FixedClasses.tex
\removelatexerror
\begin{algorithm}[H]\label{alg:fixedclass}
\caption{Standard deviation while fixing classes}
\begin{algorithmic}
\STATE $stds$ $\leftarrow$ empty list
\FOR{run $=1 \rightarrow 500$}
\STATE Select randomly $5$ classes.
\STATE $perf$ $\leftarrow$ empty list
\FOR{run $=1 \rightarrow 500$}
\STATE Select randomly $5$ shots within each class.
\STATE Generate a test set with $50$ other samples per class.
\STATE Train a LR to classify the $5$ classes from the features.
\STATE Append the LR accuracy on the test set to $perf$.
\ENDFOR
\STATE $stds$.append(standard\_deviation($perf$))
\ENDFOR
\RETURN average($stds$)
\end{algorithmic}
\end{algorithm}

%% file: Pseudo-codes/FixedShots.tex
\removelatexerror
\begin{algorithm}[H]\label{alg:fixedshot}
\caption{Standard deviation while fixing shots}
\begin{algorithmic}
\STATE $stds$ $\leftarrow$ empty list
\FOR{run $=1 \rightarrow 500$}
\STATE Select randomly $5$ shots within each class.
\STATE $perf$ $\leftarrow$ empty list
\FOR{run $=1 \rightarrow 500$}
\STATE Select randomly $5$ classes and retrieve their $5$ shots.
\STATE Generate a test set with $50$ other samples per class.
\STATE Train a LR to classify the $5$ classes from the features.
\STATE Append the LR accuracy on the test set to $perf$.
\ENDFOR
\STATE $stds$.append(standard\_deviation($perf$))
\ENDFOR
\RETURN average($stds$)
\end{algorithmic}
\end{algorithm}

%% file: Figures/supervised/fig1.tex
\begin{tikzpicture}
\begin{groupplot}[group style={group size=2 by 2, horizontal sep=2cm, vertical sep=2cm}]
	\nextgroupplot[ylabel near ticks, xlabel near ticks,
	yticklabel style={font=\Large}, xticklabel style={font=\Large},
    xlabel style={font=\LARGE, align=center}, xlabel=Number of shot per class $K$\\ (a), legend style={at={(0.95,0.05)}, anchor=south east}, ylabel style={font=\LARGE, align=center}, ymin=0, ymax=1, ylabel=\textbf{Mini-ImageNet}\\Linear correlation]
    \addplot[color=myblue, mark=x] table [x=shot, y=LR loss, col sep=comma] {csv/mini-ImageNet/WideResNet-28-10/5way-Variableshot-50query-10000run-1.csv};
    \addplot[color=myorange, mark=x] table [x=shot, y=Classes similarity, col sep=comma] {csv/mini-ImageNet/WideResNet-28-10/5way-Variableshot-50query-10000run-1.csv};
    \addplot[color=mygreen, mark=x] table [x=shot, y=N-th egv, col sep=comma] {csv/mini-ImageNet/WideResNet-28-10/5way-Variableshot-50query-10000run-1.csv};
    \addplot[color=myred, mark=x] table [x=shot, y=Davies-Bouldin score, col sep=comma, skip coords between index={0}{1}] {csv/mini-ImageNet/WideResNet-28-10/5way-Variableshot-50query-10000run-1.csv};

  
    \nextgroupplot[ylabel near ticks, xlabel near ticks,
    yticklabel style={font=\Large}, xticklabel style={font=\Large},
    xlabel style={font=\LARGE, align=center}, xlabel=Number of way $N$\\ (b), legend style={font=\Large, at={(0.98,0.98)}, anchor=north east}, ylabel style={font=\LARGE, align=center},  ymin=0, ymax=1, ylabel=Linear correlation]
    \addplot[color=myblue, mark=x] table [x=way, y=LR loss, col sep=comma] {csv/mini-ImageNet/WideResNet-28-10/Variableway-5shot-50query-10000run-1.csv};
    \addlegendentry{LR loss}
    \addplot[color=myorange, mark=x] table [x=way, y=Classes similarity, col sep=comma] {csv/mini-ImageNet/WideResNet-28-10/Variableway-5shot-50query-10000run-1.csv};
    \addlegendentry{Similarity}
    \addplot[color=mygreen, mark=x] table [x=way, y=N-th egv, col sep=comma] {csv/mini-ImageNet/WideResNet-28-10/Variableway-5shot-50query-10000run-1.csv};
    \addlegendentry{$N$-th eigenvalue}
    \addplot[color=myred, mark=x] table [x=way, y=Davies-Bouldin score, col sep=comma] {csv/mini-ImageNet/WideResNet-28-10/Variableway-5shot-50query-10000run-1.csv};
    \addlegendentry{DB score}

	\nextgroupplot[ylabel near ticks, xlabel near ticks, 
	yticklabel style={font=\Large}, xticklabel style={font=\Large},
    xlabel style={font=\LARGE, align=center}, xlabel=Number of shot per class $K$\\ (c), legend style={at={(0.95,0.05)}, anchor=south east}, ylabel style={font=\LARGE, align=center}, ymin=0, ymax=1, ylabel=\textbf{Tiered-ImageNet}\\Linear correlation]
    \addplot[color=myblue, mark=x] table [x=shot, y=LR loss, col sep=comma] {csv/tiered-ImageNet/DenseNet/5way-Variableshot-50query-10000run-1.csv};
    \addplot[color=myorange, mark=x] table [x=shot, y=Classes similarity, col sep=comma] {csv/tiered-ImageNet/DenseNet/5way-Variableshot-50query-10000run-1.csv};
    \addplot[color=mygreen, mark=x] table [x=shot, y=N-th egv, col sep=comma] {csv/tiered-ImageNet/DenseNet/5way-Variableshot-50query-10000run-1.csv};
    \addplot[color=myred, mark=x] table [x=shot, y=Davies-Bouldin score, col sep=comma, skip coords between index={0}{1}] {csv/tiered-ImageNet/DenseNet/5way-Variableshot-50query-10000run-1.csv};

  
    \nextgroupplot[ylabel near ticks, xlabel near ticks, 
    yticklabel style={font=\Large}, xticklabel style={font=\Large},
    xlabel style={font=\LARGE, align=center}, xlabel=Number of way $N$\\ (d), legend style={at={(0.95,0.05)}, anchor=south east}, ylabel style={font=\LARGE, align=center}, ymin=0, ymax=1, ylabel=Linear correlation]
    \addplot[color=myblue, mark=x] table [x=way, y=LR loss, col sep=comma] {csv/tiered-ImageNet/DenseNet/Variableway-5shot-50query-10000run-1.csv};
    \addplot[color=myorange, mark=x] table [x=way, y=Classes similarity, col sep=comma] {csv/tiered-ImageNet/DenseNet/Variableway-5shot-50query-10000run-1.csv};
    \addplot[color=mygreen, mark=x] table [x=way, y=N-th egv, col sep=comma] {csv/tiered-ImageNet/DenseNet/Variableway-5shot-50query-10000run-1.csv};
    \addplot[color=myred, mark=x] table [x=way, y=Davies-Bouldin score, col sep=comma] {csv/tiered-ImageNet/DenseNet/Variableway-5shot-50query-10000run-1.csv};
    
\end{groupplot}
\end{tikzpicture}

%% file: Figures/supervised/fig2.tex
\begin{tikzpicture}
\begin{groupplot}[group style={group size=2 by 1, horizontal sep=2cm, vertical sep=2cm}]
	\nextgroupplot[ylabel near ticks, xlabel near ticks,
	enlargelimits=false, axis on top,
	yticklabel style={font=\Large},
	 xticklabel style={font=\large, /pgf/number format/fixed, /pgf/number format/precision=2},
    xlabel style={font=\LARGE, align=center},
    xlabel=LR loss\\ (a),
    ylabel style={font=\LARGE, align=center},
    ylabel=Accuracy (\%)]
    \addplot graphics[xmin=0,xmax=0.15,ymin=25,ymax=100]{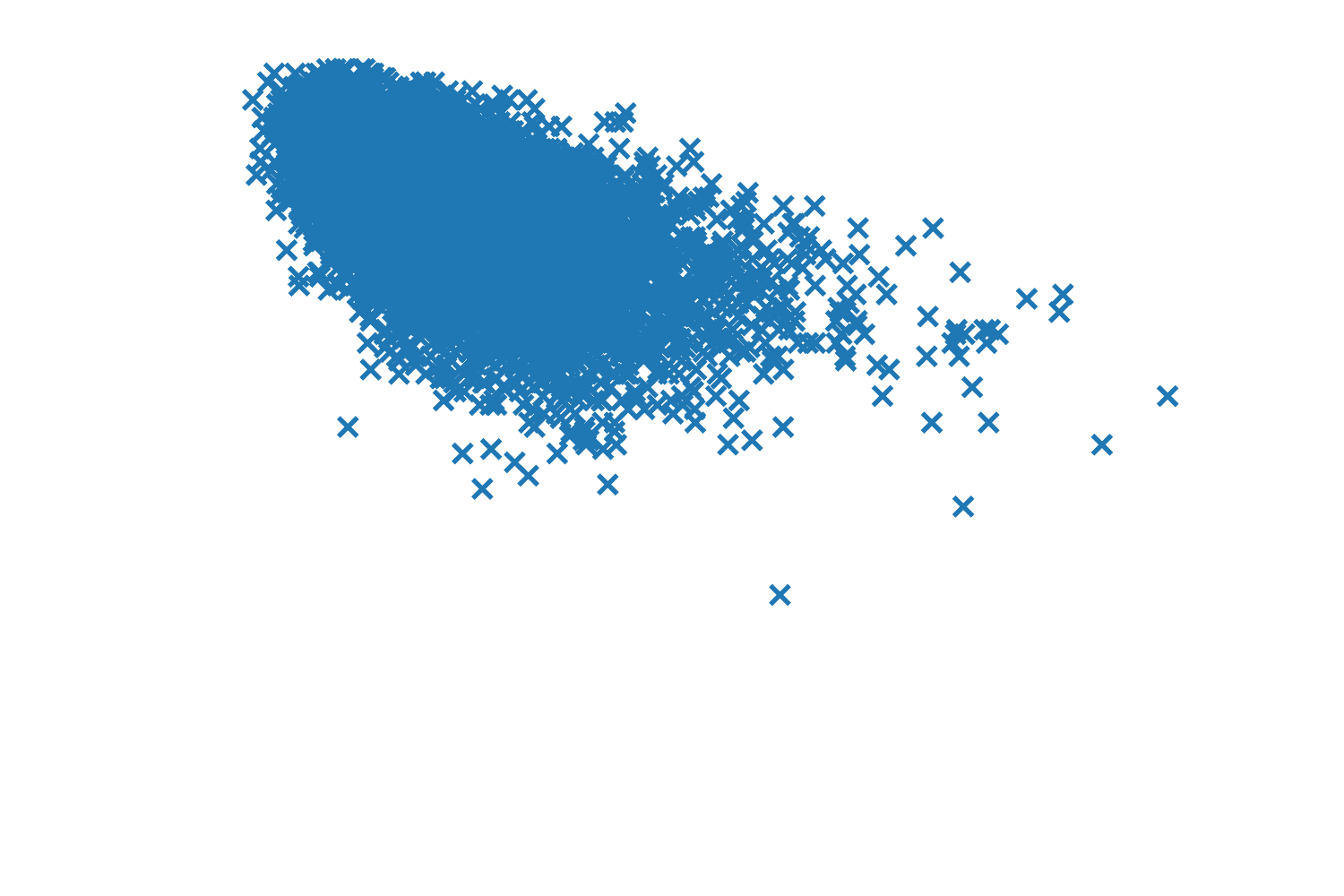};
    
	\nextgroupplot[ylabel near ticks, xlabel near ticks,
	enlargelimits=false, axis on top,
	yticklabel style={font=\Large},
	 xticklabel style={font=\large, /pgf/number format/fixed, /pgf/number format/precision=2},
    xlabel style={font=\LARGE, align=center},
    xlabel=LR loss\\ (b),
    ylabel style={font=\LARGE, align=center},
    ylabel=Accuracy (\%)]
    \addplot graphics[xmin=0,xmax=0.15,ymin=25,ymax=100]{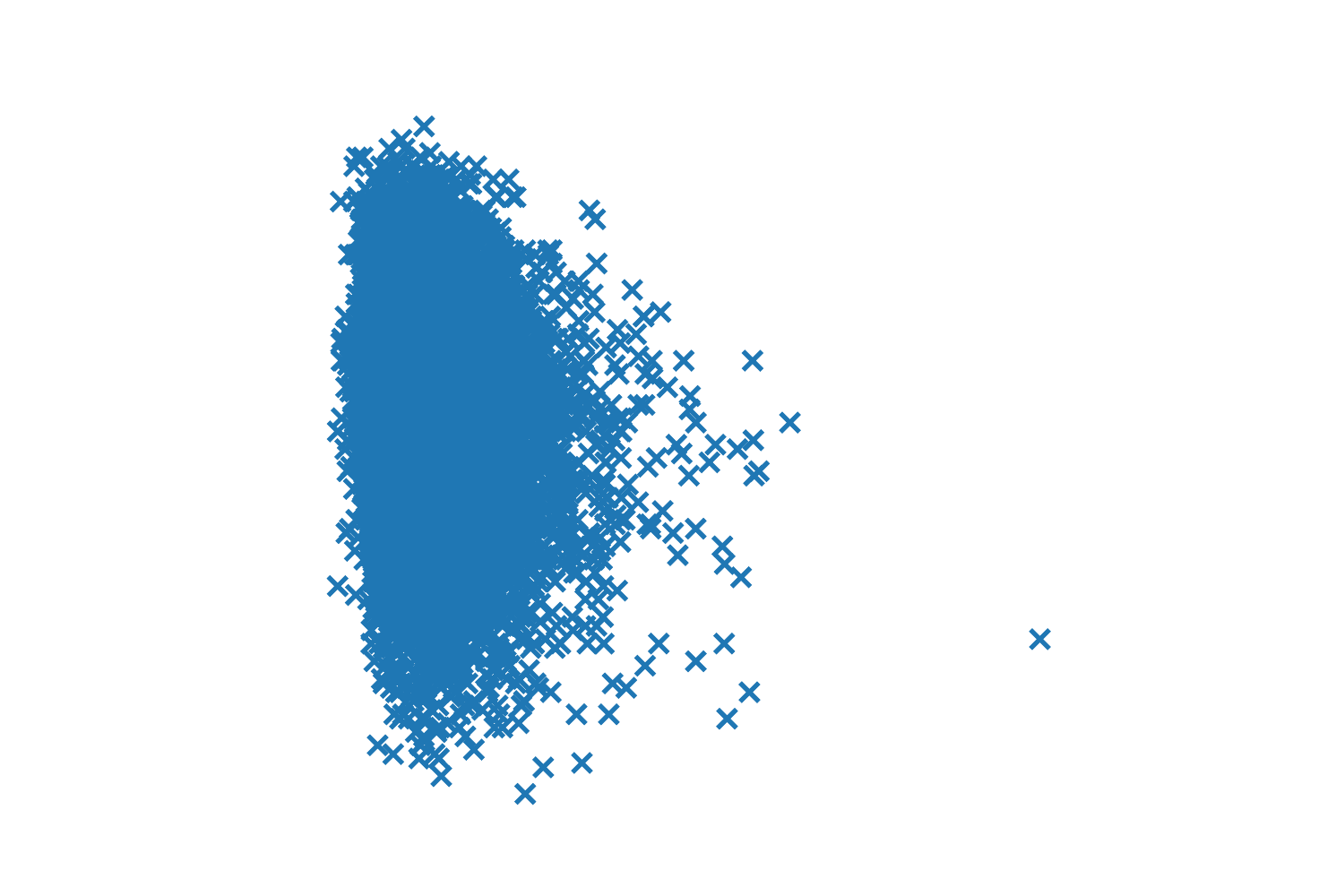};

\end{groupplot}
\end{tikzpicture}

%% file: Figures/unsupervised/fig1.tex
\begin{tikzpicture}
\begin{groupplot}[group style={group size=2 by 2, horizontal sep=2cm, vertical sep=3cm}]
	\nextgroupplot[ylabel near ticks, xlabel near ticks,
	yticklabel style={font=\Large}, xticklabel style={font=\Large},
    xlabel style={font=\LARGE, align=center}, xlabel=Number of unlabeled\\ samples per class $Q$\\(a), legend style={font=\Large, at={(0.95,0.05)}, anchor=south east}, ymin=0, ymax=1, ylabel style={font=\LARGE, align=center}, ylabel=\textbf{Mini-ImageNet}\\Linear correlation]
    \addplot[color=mygreen, mark=x] table [x=n_unsup, y=N-th egv, col sep=comma] {csv/mini-ImageNet/WideResNet-28-10/5way-5shot-50query-10000run-3.csv};
    \addplot[color=myred, mark=x] table [x=n_unsup, y=Davies-Bouldin score, col sep=comma] {csv/mini-ImageNet/WideResNet-28-10/5way-5shot-50query-10000run-3.csv};

    \nextgroupplot[ylabel near ticks, xlabel near ticks, 
    yticklabel style={font=\Large}, xticklabel style={font=\Large},
    xlabel style={font=\LARGE, align=center}, xlabel=Number of way $N$\\ \\(b), legend style={font=\Large, at={(0.98,0.98)}, anchor=north east}, ylabel style={font=\LARGE, align=center}, ymin=0, ymax=1, ylabel=Linear correlation]
    \addplot[color=mygreen, mark=x] table [x=way, y=N-th egv, col sep=comma] {csv/mini-ImageNet/WideResNet-28-10/Variableway-5shot-50query-10000run-3.csv};
    \addlegendentry{$N$-th eigenvalue}
    \addplot[color=myred, mark=x] table [x=way, y=Davies-Bouldin score, col sep=comma] {csv/mini-ImageNet/WideResNet-28-10/Variableway-5shot-50query-10000run-3.csv};
    \addlegendentry{DB score}
    
	\nextgroupplot[ylabel near ticks, xlabel near ticks, 
	yticklabel style={font=\Large}, xticklabel style={font=\Large},
    xlabel style={font=\LARGE, align=center}, xlabel=Number of unlabeled\\ samples per class $Q$\\(c), legend style={at={(0.95,0.05)}, anchor=south east}, ymin=0, ymax=1, ylabel style={font=\LARGE, align=center}, ylabel=\textbf{Tiered-ImageNet}\\Linear correlation]
    \addplot[color=mygreen, mark=x] table [x=n_unsup, y=N-th egv, col sep=comma] {csv/tiered-ImageNet/DenseNet/5way-5shot-50query-10000run-3.csv};
    \addplot[color=myred, mark=x] table [x=n_unsup, y=Davies-Bouldin score, col sep=comma] {csv/tiered-ImageNet/DenseNet/5way-5shot-50query-10000run-3.csv};

    \nextgroupplot[ylabel near ticks, xlabel near ticks, 
    yticklabel style={font=\Large}, xticklabel style={font=\Large},
    xlabel style={font=\LARGE, align=center}, xlabel=Number of way $N$\\ \\(d), legend style={at={(0.95,0.05)}, anchor=south east}, ylabel style={font=\LARGE, align=center}, ymin=0, ymax=1, ylabel=Linear correlation]
    \addplot[color=mygreen, mark=x] table [x=way, y=N-th egv, col sep=comma] {csv/tiered-ImageNet/DenseNet/Variableway-5shot-50query-10000run-3.csv};
    \addplot[color=myred, mark=x] table [x=way, y=Davies-Bouldin score, col sep=comma] {csv/tiered-ImageNet/DenseNet/Variableway-5shot-50query-10000run-3.csv};
                
\end{groupplot}
\end{tikzpicture}

%% file: Figures/semi-supervised/fig1.tex
\begin{tikzpicture}
\begin{groupplot}[group style={group size=4 by 2, horizontal sep=2cm, , vertical sep=3cm}]
	\nextgroupplot[ylabel near ticks, xlabel near ticks, 
	yticklabel style={font=\Large}, xticklabel style={font=\Large},
    xlabel style={font=\LARGE, align=center}, xlabel=Number of unlabeled\\ samples per class $Q$\\(a), legend style={font=\Large, at={(0.98,0.02)}, anchor=south east}, ymin=0, ymax=1, ylabel style={font=\LARGE, align=center}, ylabel=\textbf{Mini-ImageNet}\\Linear correlation]
    \addplot[color=myblue, mark=x] table [x=n_unsup, y=LR loss, col sep=comma] {csv/mini-ImageNet/WideResNet-28-10/5way-5shot-50query-10000run-2.csv};
    \addplot[color=myorange, mark=x] table [x=n_unsup, y=Classes similarity, col sep=comma] {csv/mini-ImageNet/WideResNet-28-10/5way-5shot-50query-10000run-2.csv};
    \addplot[color=mygreen, mark=x] table [x=n_unsup, y=N-th egv, col sep=comma] {csv/mini-ImageNet/WideResNet-28-10/5way-5shot-50query-10000run-2.csv};
    \addplot[color=myred, mark=x] table [x=n_unsup, y=Davies-Bouldin score, col sep=comma] {csv/mini-ImageNet/WideResNet-28-10/5way-5shot-50query-10000run-2.csv};
    \addplot[color=mypurple, mark=x] table [x=n_unsup, y=LR confidence, col sep=comma] {csv/mini-ImageNet/WideResNet-28-10/5way-5shot-50query-10000run-2.csv};
    
	\nextgroupplot[ylabel near ticks, xlabel near ticks, 
	yticklabel style={font=\Large}, xticklabel style={font=\Large},
    xlabel style={font=\LARGE, align=center}, xlabel=Number of shot per class $K$\\ \\(b), legend style={at={(0.95,0.05)}, anchor=south east}, ylabel style={font=\LARGE, align=center}, ymin=0, ymax=1, ylabel=Linear correlation]
    \addplot[color=myblue, mark=x] table [x=shot, y=LR loss, col sep=comma] {csv/mini-ImageNet/WideResNet-28-10/5way-Variableshot-50query-10000run-2.csv};
    \addplot[color=myorange, mark=x] table [x=shot, y=Classes similarity, col sep=comma] {csv/mini-ImageNet/WideResNet-28-10/5way-Variableshot-50query-10000run-2.csv};
    \addplot[color=mygreen, mark=x] table [x=shot, y=N-th egv, col sep=comma] {csv/mini-ImageNet/WideResNet-28-10/5way-Variableshot-50query-10000run-2.csv};
    \addplot[color=myred, mark=x] table [x=shot, y=Davies-Bouldin score, col sep=comma] {csv/mini-ImageNet/WideResNet-28-10/5way-Variableshot-50query-10000run-2.csv};
    \addplot[color=mypurple, mark=x] table [x=shot, y=LR confidence, col sep=comma] {csv/mini-ImageNet/WideResNet-28-10/5way-Variableshot-50query-10000run-2.csv};

    \nextgroupplot[ylabel near ticks, xlabel near ticks, 
    yticklabel style={font=\Large}, xticklabel style={font=\Large},
    xlabel style={font=\LARGE, align=center}, xlabel=Number of way $N$\\ \\(c), legend style={at={(0.95,0.05)}, anchor=south east}, ymin=0, ymax=1, ylabel style={font=\LARGE, align=center}, ylabel=Linear correlation]
    \addplot[color=myblue, mark=x] table [x=way, y=LR loss, col sep=comma] {csv/mini-ImageNet/WideResNet-28-10/Variableway-5shot-50query-10000run-2.csv};
    \addplot[color=myorange, mark=x] table [x=way, y=Classes similarity, col sep=comma] {csv/mini-ImageNet/WideResNet-28-10/Variableway-5shot-50query-10000run-2.csv};
    \addplot[color=mygreen, mark=x] table [x=way, y=N-th egv, col sep=comma] {csv/mini-ImageNet/WideResNet-28-10/Variableway-5shot-50query-10000run-2.csv};
    \addplot[color=myred, mark=x] table [x=way, y=Davies-Bouldin score, col sep=comma] {csv/mini-ImageNet/WideResNet-28-10/Variableway-5shot-50query-10000run-2.csv};
    \addplot[color=mypurple, mark=x] table [x=way, y=LR confidence, col sep=comma] {csv/mini-ImageNet/WideResNet-28-10/Variableway-5shot-50query-10000run-2.csv};
    
    \nextgroupplot[enlargelimits=false, axis on top, 
    ylabel near ticks, xlabel near ticks,
    yticklabel style={font=\Large}, xticklabel style={font=\Large, /pgf/number format/fixed, /pgf/number format/precision=2},
    xlabel style={font=\LARGE, align=center}, xlabel=LR confidence\\ \\(d), ylabel style={font=\LARGE, align=center}, ylabel= Accuracy (\%)]
    \addplot graphics[xmin=0.45,xmax=1,ymin=40,ymax=100]{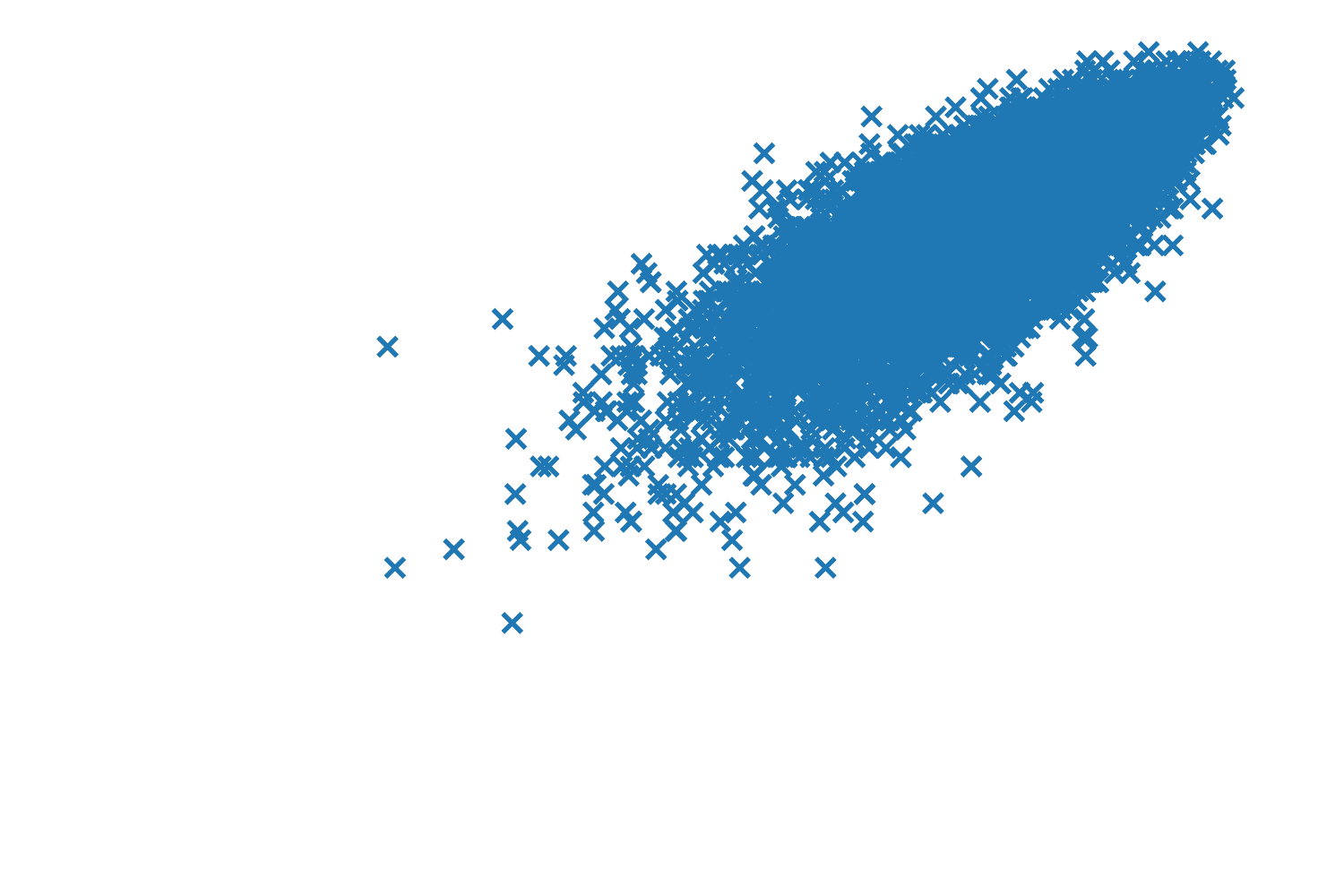};
    
	\nextgroupplot[ylabel near ticks, xlabel near ticks, 
	yticklabel style={font=\Large}, xticklabel style={font=\Large},
    xlabel style={font=\LARGE, align=center}, ymin=0, ymax=1, xlabel=Number of unlabeled\\ samples per class $Q$\\(e), legend style={font=\Large, at={(0.98,0.02)}, anchor=south east}, ylabel style={font=\LARGE, align=center}, ylabel=\textbf{Tiered-ImageNet}\\Linear correlation]
    \addplot[color=myblue, mark=x] table [x=n_unsup, y=LR loss, col sep=comma] {csv/tiered-ImageNet/DenseNet/5way-5shot-50query-10000run-2.csv};
    \addlegendentry{LR loss}
    \addplot[color=myorange, mark=x] table [x=n_unsup, y=Classes similarity, col sep=comma] {csv/tiered-ImageNet/DenseNet/5way-5shot-50query-10000run-2.csv};
    \addlegendentry{Similarity}
    \addplot[color=mygreen, mark=x] table [x=n_unsup, y=N-th egv, col sep=comma] {csv/tiered-ImageNet/DenseNet/5way-5shot-50query-10000run-2.csv};
    \addlegendentry{$N$-th eigenvalue}
    \addplot[color=myred, mark=x] table [x=n_unsup, y=Davies-Bouldin score, col sep=comma] {csv/tiered-ImageNet/DenseNet/5way-5shot-50query-10000run-2.csv};
    \addlegendentry{DB score}
    \addplot[color=mypurple, mark=x] table [x=n_unsup, y=LR confidence, col sep=comma] {csv/tiered-ImageNet/DenseNet/5way-5shot-50query-10000run-2.csv};
    \addlegendentry{LR confidence}
    
	\nextgroupplot[ylabel near ticks, xlabel near ticks, 
	yticklabel style={font=\Large}, xticklabel style={font=\Large},
    xlabel style={font=\LARGE, align=center}, ymin=0, ymax=1, xlabel=Number of shot per class $K$\\ \\(f), legend style={at={(0.95,0.05)}, anchor=south east}, ylabel style={font=\LARGE, align=center}, ylabel=Linear correlation]
    \addplot[color=myblue, mark=x] table [x=shot, y=LR loss, col sep=comma] {csv/tiered-ImageNet/DenseNet/5way-Variableshot-50query-10000run-2.csv};
    \addplot[color=myorange, mark=x] table [x=shot, y=Classes similarity, col sep=comma] {csv/tiered-ImageNet/DenseNet/5way-Variableshot-50query-10000run-2.csv};
    \addplot[color=mygreen, mark=x] table [x=shot, y=N-th egv, col sep=comma] {csv/tiered-ImageNet/DenseNet/5way-Variableshot-50query-10000run-2.csv};
    \addplot[color=myred, mark=x] table [x=shot, y=Davies-Bouldin score, col sep=comma] {csv/tiered-ImageNet/DenseNet/5way-Variableshot-50query-10000run-2.csv};
    \addplot[color=mypurple, mark=x] table [x=shot, y=LR confidence, col sep=comma] {csv/tiered-ImageNet/DenseNet/5way-Variableshot-50query-10000run-2.csv};

    \nextgroupplot[ylabel near ticks, xlabel near ticks, 
    yticklabel style={font=\Large}, xticklabel style={font=\Large},
    xlabel style={font=\LARGE, align=center}, ymin=0, ymax=1, xlabel=Number of way $N$\\ \\(g), legend style={at={(0.95,0.05)}, anchor=south east}, ylabel style={font=\LARGE, align=center}, ylabel=Linear correlation]
    \addplot[color=myblue, mark=x] table [x=way, y=LR loss, col sep=comma] {csv/tiered-ImageNet/DenseNet/Variableway-5shot-50query-10000run-2.csv};
    \addplot[color=myorange, mark=x] table [x=way, y=Classes similarity, col sep=comma] {csv/tiered-ImageNet/DenseNet/Variableway-5shot-50query-10000run-2.csv};
    \addplot[color=mygreen, mark=x] table [x=way, y=N-th egv, col sep=comma] {csv/tiered-ImageNet/DenseNet/Variableway-5shot-50query-10000run-2.csv};
    \addplot[color=myred, mark=x] table [x=way, y=Davies-Bouldin score, col sep=comma] {csv/tiered-ImageNet/DenseNet/Variableway-5shot-50query-10000run-2.csv};
    \addplot[color=mypurple, mark=x] table [x=way, y=LR confidence, col sep=comma] {csv/tiered-ImageNet/DenseNet/Variableway-5shot-50query-10000run-2.csv};

    \nextgroupplot[enlargelimits=false, axis on top, 
    ylabel near ticks, xlabel near ticks,
    yticklabel style={font=\Large}, xticklabel style={font=\Large, /pgf/number format/fixed, /pgf/number format/precision=2},
    xlabel style={font=\LARGE, align=center}, xlabel=LR confidence\\ \\(h), ylabel style={font=\LARGE, align=center}, ylabel= Accuracy (\%)]
    \addplot graphics[xmin=0.45,xmax=1,ymin=40,ymax=100]{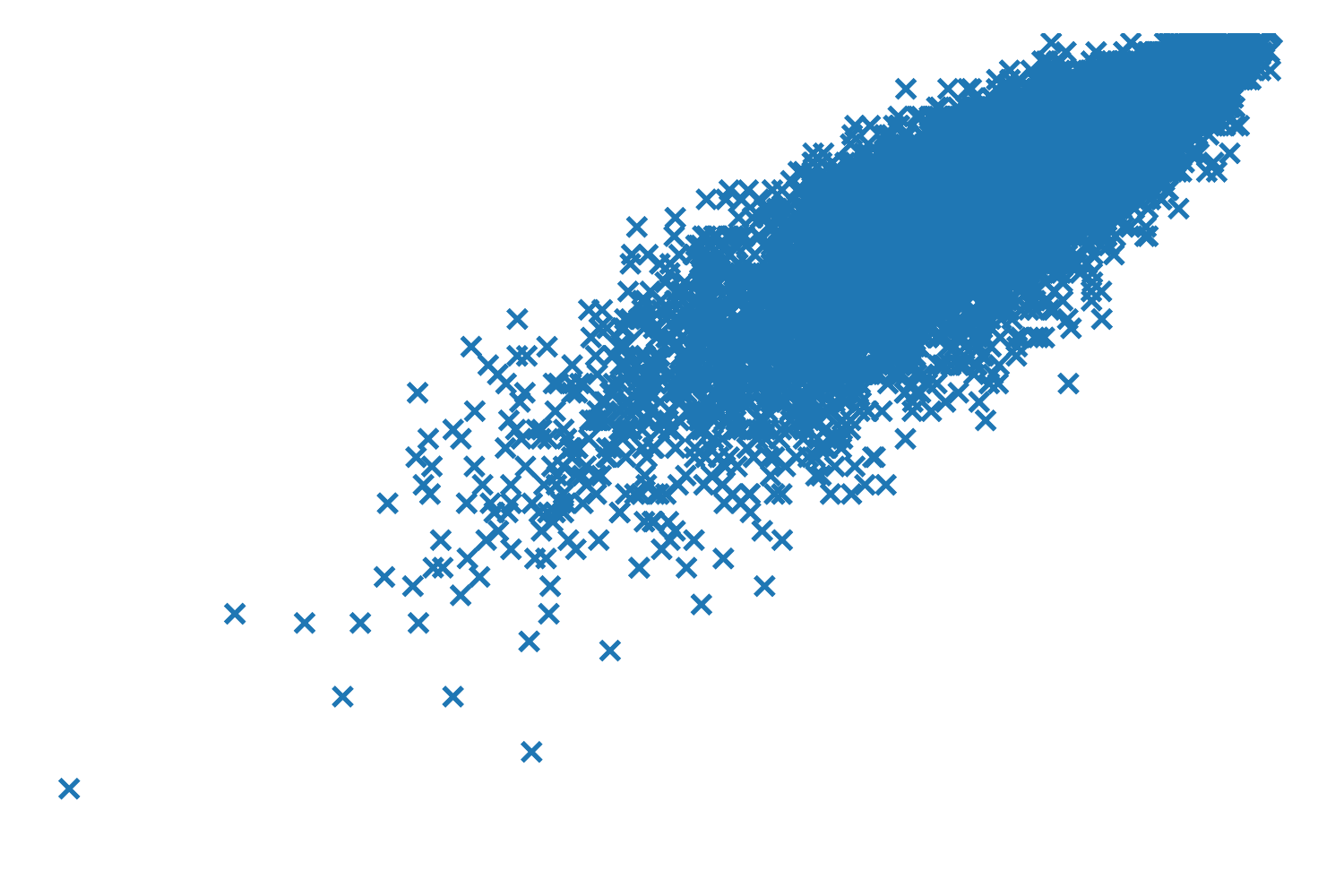};
    
\end{groupplot}
\end{tikzpicture}

%% file: Figures/prediction/conf-notations.tex
\begin{tikzpicture}[
box/.style={draw,rectangle,minimum size=1cm,text width=0.5cm},
every label/.style={text depth=depth("p")}]
\matrix (conmat) [row sep=.1cm,column sep=.1cm] {
\node (tpos) [box,
    label={[rotate=90, left=.2cm of tpos, anchor=base]left:\(\textbf{\footnotesize{hard}} \)},
    label=above:\(\textbf{\footnotesize{hard}} \),
    ] {TP};
&
\node (fneg) [box,
    label=above:\textbf{\footnotesize{easy}}]
    {FN};
\\
\node (fpos) [box,
    label={[rotate=90, left=.2cm of fpos, anchor=base]left:\( \textbf{\footnotesize{easy}} \)}]
    {FP};
&
\node (tneg) [box] {TN};
\\
};
\node [rotate=90, left=.01cm of conmat,anchor=base,text width=1.5cm,align=center] {\textbf{\footnotesize{Reality}}$\;\;\;$};
\node [above=-0.2cm of conmat] {$\;\;\;\;\;$\textbf{\footnotesize{Prediction}}};

\end{tikzpicture}

%% file: Figures/prediction/pred-super.tex
\begin{tikzpicture}[box/.style={draw,rectangle,minimum size=1cm,text width=0.5cm}]
    \begin{groupplot}[group style={group size=1 by 1, horizontal sep=2cm, vertical sep=3cm}]
	\nextgroupplot[ylabel near ticks, xlabel near ticks,
	yticklabel style={font=\Large}, xticklabel style={font=\Large},
    xlabel style={font=\LARGE, align=center}, 
    xlabel=1 -- specificity,
    legend style={at={(0.95,0.05)}, anchor=south east}, ylabel style={font=\LARGE, align=center}, ymin=-0.1, ymax=1.1, xmin=-0.1, xmax=1.1, 
    ylabel=Sensibility]
    \addplot[color=myblue, mark=x] table [x=1-specificity, y=sensibility, col sep=comma] {csv/prediction/roc-super.csv};
    \addplot+[color=myred, mark=*] plot coordinates {(0.2859,0.8094)};
\end{groupplot}

\end{tikzpicture}

%% file: Figures/prediction/conf-super.tex
\begin{tikzpicture}[
box/.style={draw,rectangle,minimum size=1cm,text width=0.5cm},
every label/.style={text depth=depth("p")}]
\matrix (conmat) [row sep=.1cm,column sep=.1cm] {
\node (tpos) [box,
    label={[rotate=90, left=.2cm of tpos, anchor=base]left:\(\textbf{\footnotesize{hard}} \)},
    label=above:\(\textbf{\footnotesize{hard}} \),
    ] {\footnotesize{$45\%$}};
&
\node (fneg) [box,
    label=above:\textbf{\footnotesize{easy}}]
    {\footnotesize{$55\%$}};
\\
\node (fpos) [box,
    label={[rotate=90, left=.2cm of fpos, anchor=base]left:\( \textbf{\footnotesize{easy}} \)}]
    {\footnotesize{$14\%$}};
&
\node (tneg) [box] {\footnotesize{$86\%$}};
\\
};
\node [rotate=90, left=.01cm of conmat,anchor=base,text width=1.5cm,align=center] {\textbf{\footnotesize{Reality}}$\;\;\;$};
\node [above=-0.2cm of conmat] {$\;\;\;\;\;$\textbf{\footnotesize{Prediction}}};

\node [below=0.5cm of conmat] {$\;$};
\end{tikzpicture}

%% file: Figures/prediction/pred-unsup.tex
\begin{tikzpicture}[box/.style={draw,rectangle,minimum size=1cm,text width=0.5cm}]
    \begin{groupplot}[group style={group size=1 by 1, horizontal sep=2cm, vertical sep=3cm}]
	\nextgroupplot[ylabel near ticks, xlabel near ticks,
	yticklabel style={font=\Large}, xticklabel style={font=\Large},
    xlabel style={font=\LARGE, align=center}, 
    xlabel=1 -- specificity,
    legend style={at={(0.95,0.05)}, anchor=south east}, ylabel style={font=\LARGE, align=center}, ymin=-0.1, ymax=1.1, xmin=-0.1, xmax=1.1, 
    ylabel=Sensibility]
    \addplot[color=myblue, mark=x] table [x=1-specificity, y=sensibility, col sep=comma] {csv/prediction/roc-unsup.csv};
    \addplot+[color=myred, mark=*] plot coordinates {(0.2983,0.8077)};
\end{groupplot}

\end{tikzpicture}

%% file: Figures/prediction/conf-unsup.tex
\begin{tikzpicture}[
box/.style={draw,rectangle,minimum size=1cm,text width=0.5cm},
every label/.style={text depth=depth("p")}]
\matrix (conmat) [row sep=.1cm,column sep=.1cm] {
\node (tpos) [box,
    label={[rotate=90, left=.2cm of tpos, anchor=base]left:\(\textbf{\footnotesize{hard}} \)},
    label=above:\(\textbf{\footnotesize{hard}} \),
    ] {\footnotesize{$95\%$}};
&
\node (fneg) [box,
    label=above:\textbf{\footnotesize{easy}}]
    {\footnotesize{$5\%$}};
\\
\node (fpos) [box,
    label={[rotate=90, left=.2cm of fpos, anchor=base]left:\( \textbf{\footnotesize{easy}} \)}]
    {\footnotesize{$64\%$}};
&
\node (tneg) [box] {\footnotesize{$36\%$}};
\\
};
\node [rotate=90, left=.01cm of conmat,anchor=base,text width=1.5cm,align=center] {\textbf{\footnotesize{Reality}}$\;\;\;$};
\node [above=-0.2cm of conmat] {$\;\;\;\;\;$\textbf{\footnotesize{Prediction}}};

\node [below=0.5cm of conmat] {$\;$};
\end{tikzpicture}

%% file: Figures/prediction/pred-semi.tex
\begin{tikzpicture}[box/.style={draw,rectangle,minimum size=1cm,text width=0.5cm}]
    \begin{groupplot}[group style={group size=1 by 1, horizontal sep=2cm, vertical sep=3cm}]
	\nextgroupplot[ylabel near ticks, xlabel near ticks,
	yticklabel style={font=\Large}, xticklabel style={font=\Large},
    xlabel style={font=\LARGE, align=center}, 
    xlabel= 1 -- specificity,
    legend style={at={(0.95,0.05)}, anchor=south east}, ylabel style={font=\LARGE, align=center}, ymin=-0.1, ymax=1.1, xmin=-0.1, xmax=1.1, 
    ylabel= Sensibility
    ]
    \addplot[color=myblue, mark=x] table [x=1-specificity, y=sensibility, col sep=comma] {csv/prediction/roc-semi.csv};
    \addplot+[color=myred, mark=*] plot coordinates {(0.161,0.806)};
\end{groupplot}

\end{tikzpicture}

%% file: Figures/prediction/conf-semi.tex
\begin{tikzpicture}[
box/.style={draw,rectangle,minimum size=1cm,text width=0.5cm},
every label/.style={text depth=depth("p")}]
\matrix (conmat) [row sep=.1cm,column sep=.1cm] {
\node (tpos) [box, align=left,
    label={[rotate=90, left=.2cm of tpos, anchor=base]left:\(\textbf{\footnotesize{hard}} \)},
    label=above:\(\textbf{\footnotesize{hard}} \),
    ] {\footnotesize{$76\%$}};
&
\node (fneg) [box, align=left,
    label=above:\textbf{\footnotesize{easy}}]
    {\footnotesize{$24\%$}};
\\
\node (fpos) [box, align=left,
    label={[rotate=90, left=.2cm of fpos, anchor=base]left:\( \textbf{\footnotesize{easy}} \)}]
    {\footnotesize{$18\%$}};
&
\node (tneg) [box, align=left] {\footnotesize{$82\%$}};
\\
};
\node [rotate=90, left=.01cm of conmat,anchor=base,text width=1.5cm,align=center] {\textbf{\footnotesize{Reality}}$\;\;\;$};
\node [above=-0.2cm of conmat] {$\;\;\;\;\;$\textbf{\footnotesize{Prediction}}};

\node [below=0.5cm of conmat] {$\;$};
\end{tikzpicture}

%% file: Figures/unbalanced.tex
\begin{tikzpicture}
\begin{groupplot}[group style={group size=2 by 2, horizontal sep=2cm, vertical sep=3cm}]

    \nextgroupplot[ylabel near ticks, xlabel near ticks,
	yticklabel style={font=\Large}, xticklabel style={font=\Large},
    xlabel style={font=\LARGE, align=center},
    xlabel=Percentage of samples in a class\\with respect to the other classes\\ (a),
    legend style={font=\Large, at={(0.98,0.02)}, anchor=south east},
    ylabel style={font=\LARGE, align=center}, ymin=0, ymax=1,
    ylabel=Linear correlation]
    
    \addplot[color=myblue, mark=x] table [x=unbalanced, y=LR loss, col sep=comma] {csv/mini-ImageNet/WideResNet-28-10/2way-5shot-50query-10000run-unbalance-2.csv};
    \addlegendentry{LR loss}
    \addplot[color=myorange, mark=x] table [x=unbalanced, y=Classes similarity, col sep=comma] {csv/mini-ImageNet/WideResNet-28-10/2way-5shot-50query-10000run-unbalance-2.csv};
    \addlegendentry{Similarity}
    \addplot[color=mygreen, mark=x] table [x=unbalanced, y=N-th egv, col sep=comma] {csv/mini-ImageNet/WideResNet-28-10/2way-5shot-50query-10000run-unbalance-2.csv};
    \addlegendentry{$N$-th eigenvalue}
    \addplot[color=myred, mark=x] table [x=unbalanced, y=Davies-Bouldin score, col sep=comma] {csv/mini-ImageNet/WideResNet-28-10/2way-5shot-50query-10000run-unbalance-2.csv};
    \addlegendentry{DB score}
    \addplot[color=mypurple, mark=x] table [x=unbalanced, y=LR confidence, col sep=comma] {csv/mini-ImageNet/WideResNet-28-10/2way-5shot-50query-10000run-unbalance-2.csv};
    \addlegendentry{LR confidence}
    
    \nextgroupplot[ylabel near ticks, xlabel near ticks,
	yticklabel style={font=\Large}, xticklabel style={font=\Large},
    xlabel style={font=\LARGE, align=center},
    xlabel=Percentage of samples in a class\\with respect to the other classes\\ (b), legend style={at={(0.95,0.05)}, anchor=south east}, ylabel style={font=\LARGE, align=center}, ymin=0, ymax=1,
    ylabel=Linear correlation]
    
    \addplot[color=myblue, mark=x] table [x=unbalanced, y=LR loss, col sep=comma] {csv/mini-ImageNet/WideResNet-28-10/5way-5shot-50query-10000run-unbalance-2.csv};
    \addplot[color=myorange, mark=x] table [x=unbalanced, y=Classes similarity, col sep=comma] {csv/mini-ImageNet/WideResNet-28-10/5way-5shot-50query-10000run-unbalance-2.csv};
    \addplot[color=mygreen, mark=x] table [x=unbalanced, y=N-th egv, col sep=comma] {csv/mini-ImageNet/WideResNet-28-10/5way-5shot-50query-10000run-unbalance-2.csv};
    \addplot[color=myred, mark=x] table [x=unbalanced, y=Davies-Bouldin score, col sep=comma] {csv/mini-ImageNet/WideResNet-28-10/5way-5shot-50query-10000run-unbalance-2.csv};
    \addplot[color=mypurple, mark=x] table [x=unbalanced, y=LR confidence, col sep=comma] {csv/mini-ImageNet/WideResNet-28-10/5way-5shot-50query-10000run-unbalance-2.csv};
    
    \nextgroupplot[ylabel near ticks, xlabel near ticks,
	yticklabel style={font=\Large}, xticklabel style={font=\Large},
    xlabel style={font=\LARGE, align=center},
    xlabel=Percentage of samples in a class\\with respect to the other classes\\ (c), 
    legend style={font=\Large, at={(0.98,0.02)}, anchor=south east},
    ylabel style={font=\LARGE, align=center}, ymin=0, ymax=1,
    ylabel=Linear correlation]
    
    \addplot[color=mygreen, mark=x] table [x=unbalanced, y=N-th egv, col sep=comma] {csv/mini-ImageNet/WideResNet-28-10/2way-5shot-50query-10000run-unbalance-3.csv};
    \addplot[color=myred, mark=x] table [x=unbalanced, y=Davies-Bouldin score, col sep=comma] {csv/mini-ImageNet/WideResNet-28-10/2way-5shot-50query-10000run-unbalance-3.csv};

    \nextgroupplot[ylabel near ticks, xlabel near ticks,
	yticklabel style={font=\Large}, xticklabel style={font=\Large},
    xlabel style={font=\LARGE, align=center},
    xlabel=Percentage of samples in a class\\with respect to the other classes\\ (d), legend style={at={(0.98,0.02)}, anchor=south east}, ylabel style={font=\LARGE, align=center}, ymin=0, ymax=1,
    ylabel=Linear correlation]
    
    \addplot[color=mygreen, mark=x] table [x=unbalanced, y=N-th egv, col sep=comma] {csv/mini-ImageNet/WideResNet-28-10/5way-5shot-50query-10000run-unbalance-3.csv};

    \addplot[color=myred, mark=x] table [x=unbalanced, y=Davies-Bouldin score, col sep=comma] {csv/mini-ImageNet/WideResNet-28-10/5way-5shot-50query-10000run-unbalance-3.csv};

\end{groupplot}
\end{tikzpicture}

%% file: Figures/egv.tex
\begin{tikzpicture}
\begin{groupplot}[group style={group size=2 by 1, horizontal sep=2cm, vertical sep=2cm}]
	\nextgroupplot[ylabel near ticks, xlabel near ticks,
	yticklabel style={font=\Large}, xticklabel style={font=\Large},
    xlabel style={font=\LARGE, align=center}, xlabel=Number of way $N$\\ (a), legend style={at={(0.95,0.05)}, anchor=south east}, ylabel style={font=\LARGE, align=center}, ymin=0, ymax=1, ylabel=Linear correlation,
    legend style={font=\Large, at={(0.98,0.98)}, anchor=north east},
    every node near coord/.append style={font=\Large}]

    \addplot[color=mygreen, mark=x] table [x=way, y=N-way egv, col sep=comma] {csv/mini-ImageNet/WideResNet-28-10/Variableway-5shot-50query-10000run-eigen-2.csv};
    \addlegendentry{$N$-th eigenvalue}
    
    \addplot[color=black, mark=x,
    visualization depends on={value \thisrow{Best egv}\as\label}, nodes near coords=\label] table [
    x=way, y=Best correlation, col sep=comma] {csv/mini-ImageNet/WideResNet-28-10/Variableway-5shot-50query-10000run-eigen-2.csv};
    \addlegendentry{Best eigenvalue}

    \nextgroupplot[ylabel near ticks, xlabel near ticks,
    yticklabel style={font=\Large}, xticklabel style={font=\Large},
    xlabel style={font=\LARGE, align=center}, xlabel=Number of way $N$\\ (b), legend style={font=\Large, at={(0.98,0.98)}, anchor=north east}, ylabel style={font=\LARGE, align=center},  ymin=0, ymax=1, ylabel=Linear correlation,
    every node near coord/.append style={font=\Large}]

    \addplot[color=mygreen, mark=x] table [x=way, y=N-way egv, col sep=comma] {csv/mini-ImageNet/WideResNet-28-10/Variableway-5shot-50query-10000run-eigen-3.csv};
    
    \addplot[color=black, mark=x,
    visualization depends on={value \thisrow{Best egv}\as\label}, nodes near coords=\label] table [x=way, y=Best correlation, col sep=comma] {csv/mini-ImageNet/WideResNet-28-10/Variableway-5shot-50query-10000run-eigen-3.csv};

\end{groupplot}
\end{tikzpicture}

%% file: Figures/annotation/annotation.tex
\begin{tikzpicture}
\begin{groupplot}[group style={group size=1 by 1, horizontal sep=2cm, vertical sep=2cm}, unit vector ratio=1 4]
    \nextgroupplot[ylabel near ticks, xlabel near ticks,
    yticklabel style={font=\footnotesize}, xticklabel style={font=\footnotesize},
    xlabel style={font=\small, align=center}, xlabel=Number of additional labeled samples, legend style={font=\footnotesize, at={(0.01,0.98)}, anchor=north west}, ylabel style={font=\small, align=center},  ymin=78, ymax=100, ylabel=Accuracy (\%)]
    \addplot[color=mygreen, mark=x] table [x=Label, y=Choice, col sep=comma] {csv/annotation.csv};
    \addlegendentry{Chosen}
    \addplot[color=black, mark=x] table [x=Label, y=Random, col sep=comma] {csv/annotation.csv};
    \addlegendentry{At random}

\end{groupplot}
\end{tikzpicture}

%% file: Figures/accuracy.tex
\begin{tikzpicture}
\begin{groupplot}[group style={group size=2 by 2, horizontal sep=2cm, vertical sep=3cm}]
	\nextgroupplot[ylabel near ticks, xlabel near ticks, 
	yticklabel style={font=\Large},
	xticklabel style={font=\Large},
    xlabel style={font=\LARGE, align=center}, xlabel=Number of unlabeled\\ samples per class $Q$\\(a), ymin=42, ymax=100,
    legend style={font=\Large, at={(0.98,0.02)}, anchor=south east}, ylabel style={font=\LARGE, align=center}, ylabel= ARI or Accuracy (\%)]
    \addplot[color=myblue, mark=x] table [x=unsup, y=Mini-LRTestAcc2, col sep=comma]
    {csv/accuracy/unsup.csv};
    \addplot[color=myorange, mark=x] table [x=unsup, y=Mini-kmAcc, col sep=comma] {csv/accuracy/unsup.csv};
    \addplot[color=mygreen, mark=x] table [x=unsup, y=Tiered-LRTestAcc2, col sep=comma] {csv/accuracy/unsup.csv};
    \addplot[color=myred, mark=x] table [x=unsup, y=Tiered-kmAcc, col sep=comma] {csv/accuracy/unsup.csv};
    
	\nextgroupplot[ylabel near ticks, xlabel near ticks, 
	yticklabel style={font=\Large}, xticklabel style={font=\Large},
	ymin=42, ymax=100,
    xlabel style={font=\LARGE, align=center}, xlabel=Number of shot per class $K$\\ \\(b), legend style={font=\Large, at={(0.98,0.02)}, anchor=south east}, ylabel style={font=\LARGE, align=center}, ylabel=ARI or Accuracy (\%)]
    
    \addplot[color=pink, mark=x] table [x=shot, y=Mini-LRTestAcc1, col sep=comma] {csv/accuracy/shot.csv};
    \addlegendentry{Mini-LR};
    
    \addplot[color=brown, mark=x] table [x=shot, y=Tiered-LRTestAcc1, col sep=comma] {csv/accuracy/shot.csv};
    \addlegendentry{Tiered-LR};
    
    \addplot[color=myblue, mark=x] table [x=shot, y=Mini-LRTestAcc2, col sep=comma] {csv/accuracy/shot.csv};
    \addlegendentry{Mini-adapted LR}

    \addplot[color=mygreen, mark=x] table [x=shot, y=Tiered-LRTestAcc2, col sep=comma] {csv/accuracy/shot.csv};
    \addlegendentry{Tiered-adapted LR}
    
    \addlegendimage{color=myorange, mark=x}
    \addlegendentry{Mini-$N$-means}
    
    \addlegendimage{color=myred, mark=x}
    \addlegendentry{Tiered-$N$-means}

    \nextgroupplot[ylabel near ticks, xlabel near ticks,
    ymin=42, ymax=100,
    yticklabel style={font=\Large}, xticklabel style={font=\Large},
    xlabel style={font=\LARGE, align=center}, xlabel=Number of way $N$\\ (c), legend style={at={(0.98,0.98)}, anchor=north east}, ylabel style={font=\LARGE, align=center}, ylabel=ARI or Accuracy (\%)]
    
    \addplot[color=myorange, mark=x] table [x=way, y=Mini-kmAcc, col sep=comma] {csv/accuracy/way.csv};
    
    \addplot[color=myred, mark=x] table [x=way, y=Tiered-kmAcc, col sep=comma] {csv/accuracy/way.csv};
    
    \addplot[color=pink, mark=x] table [x=way, y=Mini-LRTestAcc1, col sep=comma] {csv/accuracy/way.csv};
    
    \addplot[color=brown, mark=x] table [x=way, y=Tiered-LRTestAcc1, col sep=comma] {csv/accuracy/way.csv};
    
    \addplot[color=myblue, mark=x] table [x=way, y=Mini-LRTestAcc2, col sep=comma] {csv/accuracy/way.csv};
    
    \addplot[color=mygreen, mark=x] table [x=way, y=Tiered-LRTestAcc2, col sep=comma] {csv/accuracy/way.csv};
    
\end{groupplot}
\end{tikzpicture}

%% file: Figures/knn.tex
\begin{tikzpicture}
\begin{groupplot}[group style={group size=2 by 1, horizontal sep=2cm, , vertical sep=3cm}]
	\nextgroupplot[ylabel near ticks, xlabel near ticks, 
	yticklabel style={font=\Large}, xticklabel style={font=\Large},
    xlabel style={font=\LARGE, align=center},
    xlabel=Number of neighbors $k$\\(a),
    legend style={font=\Large, at={(0.98,0.98)}, anchor=north east}, ymin=0, ymax=1.1, ylabel style={font=\LARGE, align=center}, ylabel=Linear correlation]
    \addplot[color=myblue, mark=x] table [x=knn, y=LR loss, col sep=comma] {csv/mini-ImageNet/WideResNet-28-10/5way-5shot-50query-10000run-knn-2.csv};
    \addlegendentry{LR loss}
    \addplot[color=myorange, mark=x] table [x=knn, y=Classes similarity, col sep=comma] {csv/mini-ImageNet/WideResNet-28-10/5way-5shot-50query-10000run-knn-2.csv};
    \addlegendentry{Similarity}
    \addplot[color=mypurple, mark=x] table [x=knn, y=LR confidence, col sep=comma] {csv/mini-ImageNet/WideResNet-28-10/5way-5shot-50query-10000run-knn-2.csv};
    \addplot[color=mygreen, mark=x] table [x=knn, y=N-th egv, col sep=comma] {csv/mini-ImageNet/WideResNet-28-10/5way-5shot-50query-10000run-knn-2.csv};
    \addplot[color=myred, mark=x] table [x=knn, y=Davies-Bouldin score, col sep=comma] {csv/mini-ImageNet/WideResNet-28-10/5way-5shot-50query-10000run-knn-2.csv};
    
    \nextgroupplot[ylabel near ticks, xlabel near ticks, 
    yticklabel style={font=\Large}, xticklabel style={font=\Large},
    xlabel style={font=\LARGE, align=center},
    xlabel=Number of neighbors $k$\\(b),
    legend style={font=\Large, at={(0.98,0.98)}, anchor=north east}, ylabel style={font=\LARGE, align=center}, ymin=0, ymax=1.1, ylabel=Linear correlation]
    
    \addlegendimage{color=mypurple, mark=x}
    \addlegendentry{LR confidence}
    
    \addplot[color=mygreen, mark=x] table [x=knn, y=N-th egv, col sep=comma] {csv/mini-ImageNet/WideResNet-28-10/5way-5shot-50query-10000run-knn-3.csv};
    \addlegendentry{$N$-th eigenvalue}
    \addplot[color=myred, mark=x] table [x=knn, y=Davies-Bouldin score, col sep=comma] {csv/mini-ImageNet/WideResNet-28-10/5way-5shot-50query-10000run-knn-3.csv};
    \addlegendentry{DB score}
    
\end{groupplot}
\end{tikzpicture}